\pdfoutput=1

\documentclass[11pt]{article}

\usepackage{colortbl}
\usepackage[table]{xcolor}
\usepackage[final]{acl} 
\usepackage{times}
\usepackage{latexsym}
\usepackage[T1]{fontenc}
\usepackage[utf8]{inputenc}
\usepackage{microtype}
\usepackage{inconsolata}
\usepackage{setspace}

\usepackage{algorithm}
\usepackage{algorithmic}
\usepackage{graphicx}
\usepackage{amsmath} 
\usepackage{amssymb}
\usepackage{bbm}

\usepackage{pifont}
\usepackage{soul}
\usepackage{tabularx}
\usepackage{xspace}
\usepackage{makecell}
\usepackage{tipa}
\usepackage{booktabs}
\usepackage{multirow}
\usepackage{caption}
\usepackage{arydshln}
\usepackage{tcolorbox}
\usepackage{adjustbox}
\usepackage{enumitem}
\usepackage{fontawesome5}

\tcbuselibrary{skins} 

\usepackage[breaklinks]{hyperref}

\newcommand{\cmark}{\textcolor{green!60!black}{\ding{51}}} 
\newcommand{\xmark}{\textcolor{red}{\ding{55}}}

\definecolor{azure}{rgb}{0.94, 1.0, 1.0}
\definecolor{lightgray}{rgb}{0.9, 0.9, 0.9}
\definecolor{lightred}{rgb}{1.0, 0.94, 0.94}
\definecolor{decovecblueback}{RGB}{240, 248, 255} 
\definecolor{decovecblueframe}{RGB}{70, 130, 180}  
\definecolor{commentboxbg}{RGB}{255, 255, 255}     

\newcommand{\cc}[1]{\cellcolor{azure}#1}

\newtcolorbox{custombox}[1][]{
    enhanced, 
    attach boxed title to top center={yshift=-3mm}, 
    colback=decovecblueback, 
    colframe=decovecblueframe, 
    coltitle=white, 
    title=\textbf{#1}, 
    fonttitle=\bfseries, 
    boxed title style={
        colback=decovecblueframe, 
        sharp corners, 
        frame hidden, 
    },
    sharp corners=south, 
    arc=3mm, 
    drop shadow, 
    boxrule=0.5mm, 
    top=1.5em, 
}

\newtcolorbox{promptbox}[1][]{
    enhanced, 
    attach boxed title to top center={yshift=-3mm}, 
    colback=decovecblueback, 
    colframe=decovecblueframe, 
    coltitle=white, 
    title=\textbf{#1}, 
    fonttitle=\bfseries, 
    boxed title style={
        colback=decovecblueframe, 
        sharp corners, 
        frame hidden, 
    },
    sharp corners=south, 
    arc=3mm, 
    drop shadow, 
    boxrule=0.5mm, 
    top=1.5em, 
}

\title{DeCoVec: Building Decoding Space based Task Vector for Large Language Models via In-Context Learning}

\author{
    Feiyang Li ,
    Yile Wang\thanks{Corresponding to \textit{wangyile@szu.edu.cn}.} \\ 
    College of Computer Science and Software Engineering, Shenzhen University
}

\begin{document}
\maketitle

\begin{abstract}
Task vectors, representing directions in model or activation spaces that encode task-specific behaviors, have emerged as a promising tool for steering large language models (LLMs). However, existing approaches typically require fine-tuning or invasive manipulation of internal states, limiting their flexibility and scalability. We propose \textsc{DeCoVec} (Decoding Space based Task Vector), a training-free and non-invasive framework that constructs task vectors directly in the \textit{decoding space} by leveraging in-context learning (ICL). Specifically, \textsc{DeCoVec} captures the task essence as the difference between the output logit distributions of few-shot and zero-shot prompts, then steers generation by injecting this vector into the decoding process. Experiments across seven LLMs (0.5B--9B) on TruthfulQA, Math-500, and AQUA-RAT show that \textsc{DeCoVec} consistently outperforms standard few-shot baselines, with gains up to +5.50 average accuracy. Further analysis demonstrates that \textsc{DeCoVec} effectively suppresses generation degeneration and logical flaws while exhibiting strong robustness to demonstration ordering, all without incurring additional input token costs. Our method offers a training-free and non-invasive solution for LLM steering without requiring weight updates or auxiliary models. The code is released at \url{https://github.com/szu-tera/DeCoVec.git}.

\end{abstract}

\section{Introduction}
\label{sec:introduction}

Task vectors~\cite{ilharco2023editing,liu_-IncontextVectors_2024-ICV,yang2025task}, representing directions in the high-dimensional space that encode specific skills or behaviors, have garnered significant attention with the development of large language models (LLMs;~\citealp{brown2020language,openaichatgpt,achiam2023gpt}). By adjusting model parameters or internal states along these vectors, one can effectively steer the model's output to adapt to specific tasks. This concept not only serves as a cost-effective alternative to parameter-efficient fine-tuning~\cite{pmlr-v97-houlsby19a,li-liang-2021-prefix,hu_lora_2021} but also offers a unique lens for interpreting the internal mechanisms of LLMs and achieving controllable generation~\cite{hendel-etal-2023-context,yang2025task}.

\begin{figure}[t]
    \centering
    \includegraphics[width=\columnwidth, trim=0cm 6.4cm 0cm 0cm, clip]{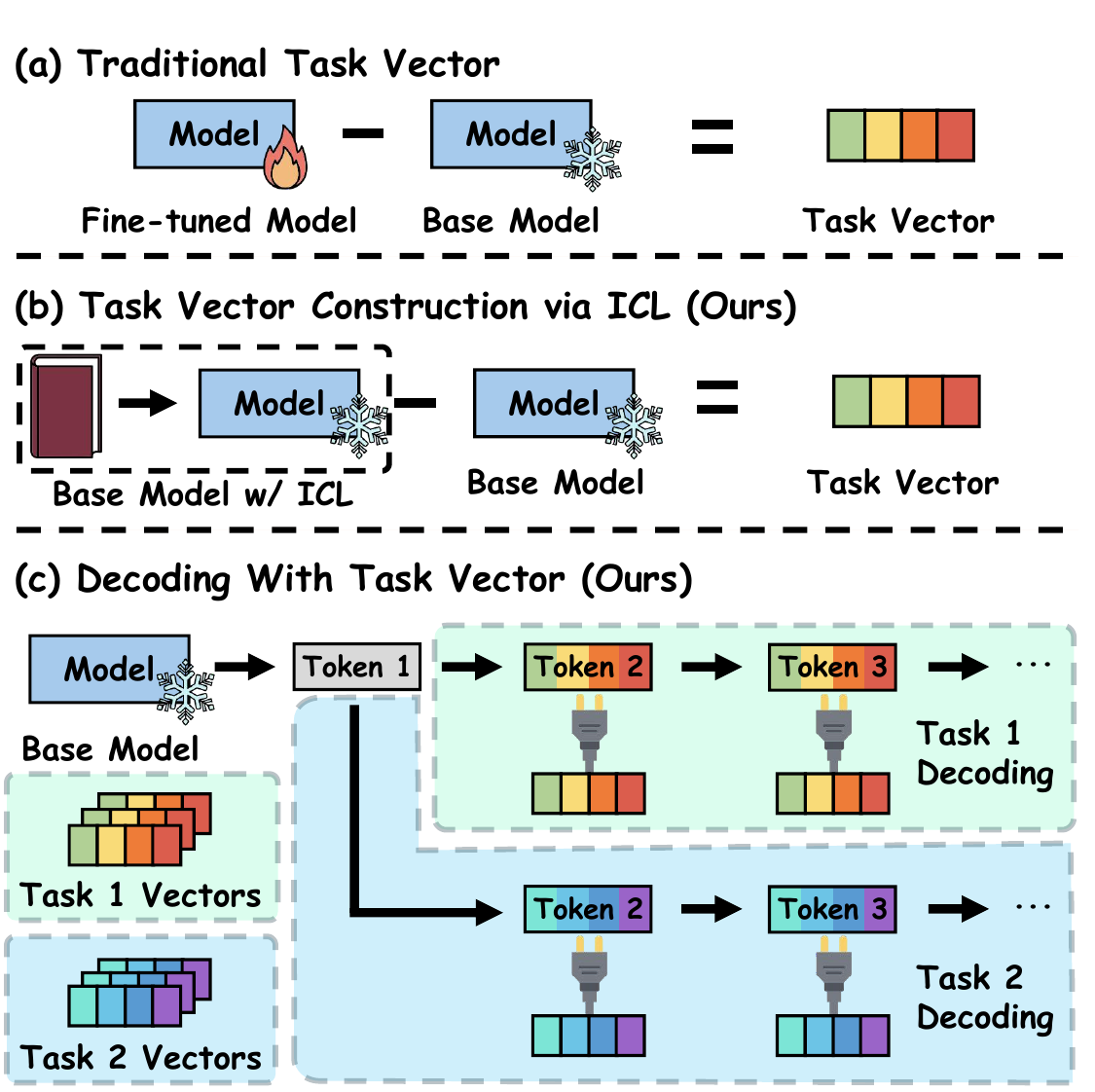} 
    \caption{
        \textbf{Schematic of traditional methods vs. ours.}
        (a) Traditional task vector in model space. 
        (b)-(c) Our \textsc{DeCoVec} via in-context learning in decoding space.
    }
    \label{fig:intro}
\end{figure}

Current task vector methods primarily operate in model spaces, either in model weights or activations, as shown in Figure~\ref{fig:intro}(a). These approaches typically require a full fine-tuning process per task to derive the vector~\cite{ilharco2023editing, huang-etal-2024-chat-chatvector}, or involve invasive intervention into the model's internal hidden states through complex optimization or auxiliary training~\cite{hendel-etal-2023-context, liu_-IncontextVectors_2024-ICV}. Their high computational cost and structural invasiveness limit flexibility and scalability in real-world deployment.

\begin{table*}[t!]
\centering
\resizebox{1\linewidth}{!}{
\begin{tabular}{lcccc}
\toprule
\textbf{Types of Task Vector}  & \textbf{Non-Invasive} & \textbf{Training-Free}& \textbf{Operational Space} & \textbf{Vector Representation} \\ 
\midrule
Arithmetical Task Vector \citep{ilharco2023editing} & \multirow{4}{*}{\xmark} & \multirow{4}{*}{\xmark} & \multirow{4}{*}{Model Weights} & \multirow{4}{*}{$ \Delta \theta$} \\
Chat Vector \citep{huang-etal-2024-chat-chatvector} & & & & \\
Emotional Vector \citep{kalyan2024emotion} & & & & \\
Contrastive Vector \citep{fierro2025steering_constructvector} & & & & \\ 
\midrule
Rule Task Vector \citep{hendel-etal-2023-context}  & \multirow{4}{*}{\xmark} & \multirow{4}{*}{\cmark} & \multirow{4}{*}{Activation States}& \multirow{4}{*}{$ \Delta h$} \\
Function Vector \citep{Todd2023FunctionVI_functionvectors} & & & & \\
In-Context Vector \citep{liu_-IncontextVectors_2024-ICV} & & & & \\
Visual Task Vector \citep{hojel2024finding_visualtaskvectors} & & & & \\ 
\midrule
\textbf{\textsc{DeCoVec} (Ours)}  & \cmark & \cmark & \textbf{Decoding Logits}& \textbf{$\mathbf{\Delta z}$} \\ 
\bottomrule
\end{tabular}}
\caption{\textbf{Comparison of task vector related works.} $\Delta\theta$: difference in model weights. $\Delta h$: difference in internal hidden states. $\Delta {\rm z}$: difference in output logits. 
Previous work mainly uses task vectors for LLMs interpretability and steering.  
Our non-invasive and training-free method operates in the decoding space for improving task performance.}
\label{tab:task_vector_comparison}
\end{table*}

On the other hand, decoding-based strategies are progressively being utilized to steer LLMs. For example, modulating output probability distributions can effectively improve generation quality~\citep{li-etal-2023-contrastive,chuang2024dola} or adapt models to downstream tasks~\citep{Liu2024TuningLM,hu2025distributionaligned,Shi2023TrustingYE}, showing that the decoding space can serve as a \textbf{semantic-rich and direct} interface for controlling model behaviors without accessing internal parameters.

In this work, we propose \textsc{DeCoVec}, a training-free and non-invasive framework for constructing task vectors directly in the decoding space. As shown in Figure~\ref{fig:intro}(b)-(c), we leverage the emergent in-context learning (ICL; \citealp{Brown2020LanguageMA}) capability of LLMs, capturing task essence by contrasting the logit distributions of zero-shot and few-shot contexts without gradient updates. By defining the task vector in the output logit space rather than internal layers, \textsc{DeCoVec} steers generation in a transparent and controllable manner. This design makes \textsc{DeCoVec} a lightweight, plug-and-play solution that decouples task guidance from the internal model architecture.

We extensively evaluate a diverse suite of 0.5B$\sim$9B LLMs~\cite{yang_qwen2_2024,ai_yi_2025,touvron_llama_2023-llama2,grattafiori2024llama,team_gemma_2024} on both knowledge-intensive TruthfulQA~\cite{lin-etal-2022-truthfulqa} and reasoning tasks including Math-500~\cite{lightman_let_2024_math500} and AQUA-RAT~\cite{ling-etal-2017-program-aquarat}. Results show that \textsc{DeCoVec} consistently achieves performance gains, outperforming few-shot baselines with different demonstration selection strategies. In-depth analysis reveals that the extracted vectors encode high-level task semantics rather than surface-level patterns. Error analysis on mathematical reasoning tasks indicates that \textsc{DeCoVec} effectively suppresses logical flaws and generation degeneration, steering models toward more rigorous problem-solving strategies. These findings offer new insights for efficiently steering LLM behavior.
\section{Related Work}

\noindent\textbf{Task Vector.} The concept of task vector aligns with the broader goal of parameter-efficient fine-tuning, offering a cost-effective way to interpret or adapt models by modifying weights via algebraic operations or injecting low-rank modules~\citep{hu_lora_2021}. Current methods can be categorized into two types according to the operational space:

1) \textit{Weight-space based Task Vector.} This type of method directly manipulates model parameters to alter behavior. \citet{ilharco2023editing} introduces the arithmetic task vector, derived by subtracting the weights of a pre-trained model from a fine-tuned one to isolate task properties. Subsequent works have extended this foundational paradigm to various specialized domains. For example, the chat vector \citep{huang-etal-2024-chat-chatvector} transfers conversational capabilities to base models. The emotional vector \citep{kalyan2024emotion} is constructed to steer the desired sentiment of generation. The contrastive vector \citep{fierro2025steering_constructvector} isolates specific behavioral patterns by contrasting weights between models tuned to opposing objectives.

\begin{figure*}[t]
  \centering
  \includegraphics[width=1\textwidth]{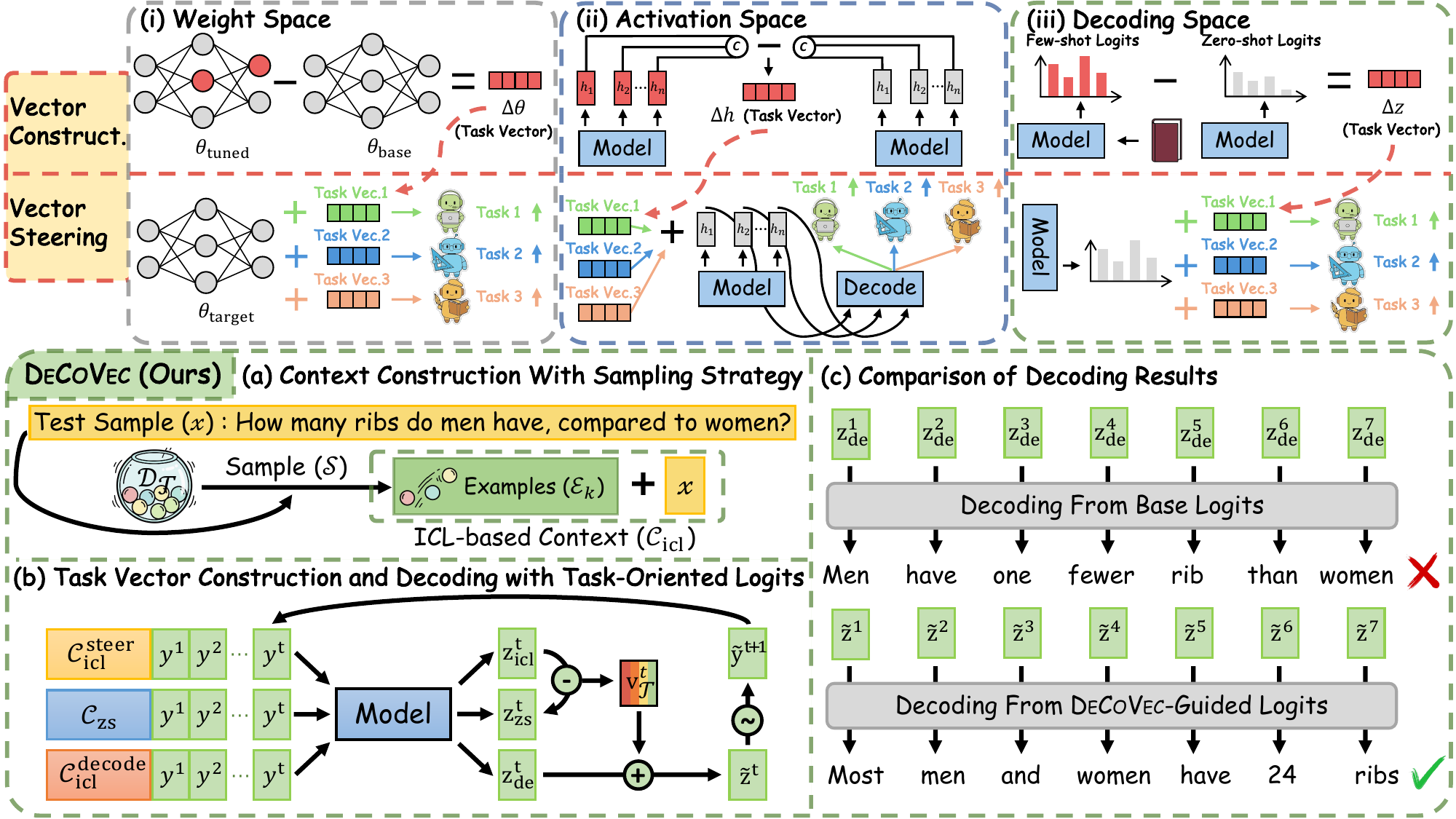}
  \caption{\textbf{Overview of task vector methods across spaces and our \textsc{DeCoVec}.} \textbf{Top: Comparison of task vector construction and steering.} (i) Weight space vector. (ii) Activation space vector. (iii) Our proposed \textsc{DeCoVec} in the decoding space. \textbf{Bottom: Illustration of \textsc{DeCoVec} pipeline.} (a) Context construction. (b) Task vector building and steering. (c) Resulting improvement in decoding output (correct \cmark vs. incorrect \xmark).}
  \label{fig:decovec_pipeline}
\end{figure*}

2) \textit{Activation-space based Task Vector.} Shifting from static parameters to dynamic representations, these types of methods intervene in internal hidden states during inference. \citet{hendel-etal-2023-context} identifies rule task vectors demonstrating that compressing demonstrations into a single vector within the activation space can replicate specific functions without explicit fine-tuning. \citet{liu_-IncontextVectors_2024-ICV} proposes the in-context vector to steer model behavior by shifting hidden states. Other explorations include the function vector~\citep{Todd2023FunctionVI_functionvectors}, which locates causal attention heads for specific tasks, and the visual task vector \citep{hojel2024finding_visualtaskvectors} specifically adapted for multimodal contexts.

\noindent\textbf{In-Context Learning of LLMs.}
ICL empowers LLMs to perform novel tasks by conditioning on input-output examples without explicit parameter updates~\citep{Brown2020LanguageMA}.
Theoretical frameworks suggest that ICL implicitly implements optimization algorithms like gradient descent \citep{Dai2023WhyCG, Bai2023TransformersAS} or serves as semantic anchors from an information flow perspective \citep{wang-etal-2023-label}. 
\citet{hendel-etal-2023-context} interprets the internal mechanism of ICL through identifying rule task vectors, while our work directly aims at constructing and steering decoding-based task vectors to improve task-specific performance.

\noindent\textbf{Decoding Strategies for LLMs.}
The decoding space serves as a direct and flexible interface for steering model behaviors. Contrastive decoding \citep{li-etal-2023-contrastive} manipulates this space to mitigate degenerative behaviors by contrasting expert and amateur logits.
Subsequent research extended this paradigm to suppress hallucinations \citep{chuang2024dola} or noise \citep{Zhou2025ALWAL}.
More recently, efforts have shifted towards modeling downstream \textit{task semantics} rather than just quality control. Proxy-tuning \citep{Liu2024TuningLM} and distribution-aligned decoding \citep{hu2025distributionaligned} adjust output distributions to approximate fine-tuned models. Similarly, \citet{peng_enhancing_2025} enhanced input-label mappings via contrastive measures.
These methods often rely on auxiliary models, gradients, or complex contrastive setups. In contrast, our approach extracts task vectors purely from the decoding space of a frozen model via ICL, offering a lightweight and non-invasive alternative.

We summarize the main related works and compare them with ours in Table~\ref{tab:task_vector_comparison}.
\section{Method}
We first formalize the problem setting and the construction of task-specific contexts (\S\ref{subsec:formalization}). Then, we derive the definition of the task vector in the decoding space by contrasting the distributions of few-shot and zero-shot settings (\S\ref{subsec:steering}). Finally, we present the steering mechanism to guide the process of decoding for downstream tasks (\S\ref{subsec:decoding}).

\subsection{Formalization of Tasks and Contexts}
\label{subsec:formalization}

\noindent\textbf{Tasks.} Considering a task $\mathcal{T}$ with input-output training pairs $\mathcal{D}_\mathcal{T} = \{(x_i, y_i)\}$, we use a task-specific instruction template (or prompt schema) $\mathcal{I}_\mathcal{T}(\cdot)$ as the input for the LLM $\mathcal{M}_\theta$ parameterized by $\theta$. Given a query input $x_i \in \mathcal{D}_\mathcal{T}$, the goal is to generate the target sequence $y_i = (y_i^1, y_i^2, \dots, y_i^L)$. At each time step $t$, the model $\mathcal{M}_\theta$ maps the context to a probability distribution over the vocabulary $\mathcal{V}$ via a softmax function applied to the output logits by model function $f_\theta(\cdot) \in \mathbb{R}^{|\mathcal{V}|}$.

\noindent\textbf{Contexts.} To leverage contexts for building task vector, we consider two distinct contextual representations of the same task $\mathcal{T}$:

1) \textit{Zero-Shot Context (Task-Agnostic State).} We construct the zero-shot input $\mathcal{C}_{\rm zs}$ by wrapping the query $x$ solely with the task instruction without explicit task demonstrations, representing the model's intrinsic instruction-following capability:
\begin{equation}
    \mathcal{C}_{\rm zs} = \mathcal{I}_\mathcal{T}(\varnothing, x),
\end{equation}
where $\varnothing$ indicates an empty set of demonstrations.

2) \textit{Few-Shot ICL-based Context (Task-Aware State).} To activate the model's emergent ability for task $\mathcal{T}$ through ICL, we employ a demonstration sampler $\mathcal{S}$ to retrieve a set of $k$ exemplars $\mathcal{E}_k$ from dataset $\mathcal{D}_\mathcal{T}$ conditioned on the query input $x$:
\begin{equation}
    \mathcal{E}_k = \mathcal{S}(x, \mathcal{D}_\mathcal{T}, k),
\end{equation}
where $\mathcal{S}$ represents a generic sampling strategy, which can range from random selection to distinct retrieval methods such as KATE \cite{liu2021KATE} or BM25~\cite{robertson2009BM25}. The few-shot ICL-based $\mathcal{C}_{\rm icl}$ is then constructed by organizing these demonstrations alongside the query within the task template:
\begin{equation}
    \mathcal{C}_{\rm icl} = \mathcal{I}_\mathcal{T}(\mathcal{E}_k, x),
\end{equation}
where $\mathcal{C}_{\rm icl}$ contains input-label mappings and semantic patterns of task $\mathcal{T}$, which are absent in $\mathcal{C}_{\rm zs}$.

\noindent\textbf{Two Types of ICL Contexts.} By applying a specific sample strategy $\mathcal{S}$, we construct $\mathcal{C}_{\rm icl}$ in two scenarios. We first sample the base context during decoding, denoted as $\mathcal{C}^{\rm decode}_{\rm icl}$ in the following Eq.~\ref{eq:decoding}, serving as a few-shot baseline. 
Then we sample the context from training pairs $\mathcal{D}_\mathcal{T}$ for the task vectors used in steering, which we denote as $\mathcal{C}^{\rm steer}_{\rm icl}$ and is used in the following Eq.~\ref{eq:context_steer}.

\subsection{Building Task Vector in Decoding Space}
\label{subsec:steering}

As illustrated in the top row of Figure~\ref{fig:decovec_pipeline}, unlike prior methods that extract task vectors from model weights (i) or internal activations (ii), we propose to construct the task vector directly within the \textit{decoding space} (iii). We hypothesize that the difference between the logits generated in the few-shot ICL-based context $\mathcal{C}^{\rm steer}_{\rm icl}$ and the zero-shot context $\mathcal{C}_{\rm zs}$ encodes the distilled task-level features that can help construct the task vector $\mathbf{v}_{\mathcal{T}}$.

Given the prefix $y^{1:t}$ at decoding step $t$, we compute the logit vector $\mathbf{z}^{t}$ $\in$ $\mathbb{R}^{|\mathcal{V}|}$ using zero-shot context and sampled ICL-based  context, respectively:
\begin{align}
    \mathbf{z}_{\rm zs}^{t} &= f_\theta(\mathcal{C}_{\rm zs}, y^{1:t}), \\
    \mathbf{z}_{\rm icl}^{t} &= f_\theta(\mathcal{C}^{\rm steer}_{\rm icl}, y^{1:t}).
    \label{eq:context_steer}
\end{align}

Given the above two logits at generation step $t$, we define the decoding-space in-context learning guided task vector  (i.e., our \textsc{DeCoVec}) $\mathbf{v}_{\mathcal{T}}^{t}$ as the difference of the two:
\begin{equation}
    \mathbf{v}_{\mathcal{T}}^{t} = \mathbf{z}_{\rm icl}^{t} - \mathbf{z}_{\rm zs}^{t}.
\label{eq:steering}
\end{equation}
This operation is depicted in Figure~\ref{fig:decovec_pipeline}(b).

\subsection{Decoding with Task Oriented Logits}
\label{subsec:decoding}

We employ our \textsc{DeCoVec} $\mathbf{v}^t_{\mathcal{T}}$ to directly steer the generation trajectory of the model. The core intuition is to inject the distilled task-specific signal into the inference process.


During decoding, the steering operates token-by-token online. We first compute the intermediate logit $\mathbf{z}_{\rm de}^{t}$ according to the base context $\mathcal{C}^{\rm decode}_{\rm icl}$. Because both $\mathbf{z}_{\rm de}^{t}$ and the task vector $\mathbf{v}^t_{\mathcal{T}}$ condition on the identical generated prefix $y^{1:t}$, their vocabulary distributions remain strictly aligned at each step, inherently avoiding sequence-length mismatches across different contexts. We then compute the final task-oriented logits $\tilde{\mathbf{z}}^{t}$ by injecting \textsc{DeCoVec} $\mathbf{v}^t_{\mathcal{T}}$:
\begin{align}
    \mathbf{z}_{\rm de}^{t} &= f_\theta(\mathcal{C}^{\rm decode}_{\rm icl}, y^{1:t}),
    \label{eq:decoding}
    \\
    \tilde{\mathbf{z}}^{t} &= \mathbf{z}_{\rm de}^{t} + \lambda \cdot \mathbf{v}_{\mathcal{T}}^{t},
    \label{eq:decoding_steering}
\end{align}
where the scaling factor $\lambda>0$ controls the task signal injected into the decoding process. This steering operation, which is the core mechanism of our method, is illustrated in Figure~\ref{fig:decovec_pipeline}(b).

This formulation allows us to explicitly control the impact of the extracted task information. The base logits $\mathbf{z}_{\rm de}^{t}$ serve as a semantic anchor, ensuring that the generation remains grounded in the logic of the current input while benefiting from the directional steering provided by $\mathbf{v}_{\mathcal{T}}^{t}$. Increasing $\lambda$ directly amplifies the task-specific signals derived from the difference between contexts, thereby encouraging the model to adhere more strictly to the intended task behaviors. We analyze the impact of scaling factor $\lambda$ in Section~\ref{subsec:lambda}.

\definecolor{azure}{rgb}{0.94, 1.0, 1.0}
\definecolor{lightgray}{rgb}{0.9, 0.9, 0.9}
\definecolor{lightred}{rgb}{1.0, 0.9, 0.9} 

\begin{table*}[t!]
  \centering
  \small
  \setlength{\fboxsep}{1.5pt} 
  \setlength{\tabcolsep}{2.5pt} 
  \renewcommand{\arraystretch}{1.2}

  \newcommand{\cbox}[1]{\colorbox{azure}{#1}}

  \newcommand{\rbox}[1]{\colorbox{lightred}{#1}} 
  
  \newcommand{\up}[2]{#1 \cbox{($\uparrow$\textbf{#2})}}

  \newcommand{\down}[2]{#1 \rbox{($\downarrow$#2)}} 
  
  \newcommand{\avg}[1]{\cbox{\textbf{#1}}}

  \newcommand{\ravg}[1]{\rbox{\textbf{#1}}}

  \begin{tabular}{l l ccc c cc c}
    \toprule
     & 
    \multirow{2.5}{*}{\textbf{Methods}} &\multicolumn{3}{c}{\textbf{TruthfulQA}} &
    \multirow{2.5}{*}{\textbf{Avg. $\Delta$}} & 
    \textbf{Math-500} &
    \textbf{AQUA-RAT} &
    \multirow{2.5}{*}{\textbf{Avg. $\Delta$}} \\ 
    
    \cmidrule(lr){3-5} \cmidrule(lr){7-7} \cmidrule(lr){8-8}
    
     & & {MC1} & {MC2} & {MC3} & & {Acc.} & {Acc.} & \\
    \midrule
    
    \multirow{5.5}{*}{\rotatebox{90}{\textbf{Qwen2-0.5B}}} 
      & Zero-Shot & 25.18 & 40.78 & 19.55 & - & 1.17 & 20.87 & - \\
      \cdashline{2-9} \addlinespace[2pt]
      & Few-Shot (Rand. Std.) & \up{32.35}{1.97} & \up{47.54}{2.51} & \up{23.46}{1.38} & \avg{$\uparrow$1.95} & \down{\phantom{0}8.59}{0.39} & \up{23.23}{5.12} & \avg{$\uparrow$2.37} \\
      & Few-Shot (Rand. Ext.) & \up{32.21}{0.70} & \up{48.01}{2.55} & \up{23.73}{1.42} & \avg{$\uparrow$1.56} & \up{\phantom{0}7.03}{0.78} & \up{21.65}{3.55} & \avg{$\uparrow$2.17} \\
      & Few-Shot (KATE)       & \up{34.60}{0.42} & \up{50.43}{2.12} & \up{24.98}{1.57} & \avg{$\uparrow$1.37} & \up{\phantom{0}8.59}{3.52} & \up{18.11}{5.91} & \avg{$\uparrow$4.72} \\
      & Few-Shot (BM25)       & \up{32.35}{0.28} & \up{47.83}{2.64} & \up{23.09}{1.36} & \avg{$\uparrow$1.43} & \down{10.55}{0.39} & \up{20.87}{1.96} & \avg{$\uparrow$0.79} \\
    \midrule

    \multirow{5.5}{*}{\rotatebox{90}{\textbf{Qwen2-1.5B}}} 
      & Zero-Shot & 26.30 & 44.70 & 20.78 & - & 0.78 & 8.66 & - \\
      \cdashline{2-9} \addlinespace[2pt]
      & Few-Shot (Rand. Std.) & \up{33.47}{0.99} & \up{49.91}{0.48} & \up{25.02}{0.46} & \avg{$\uparrow$0.64} & \up{17.58}{0.78} & \up{24.80}{4.33} & \avg{$\uparrow$2.56} \\
      & Few-Shot (Rand. Ext.) & \up{42.76}{1.12} & \up{60.33}{2.27} & \up{32.24}{2.03} & \avg{$\uparrow$1.81} & \up{17.97}{0.39} & \up{29.13}{1.58} & \avg{$\uparrow$0.99} \\
      & Few-Shot (KATE)       & \up{37.41}{0.99} & \up{53.89}{0.62} & \up{27.47}{0.84} & \avg{$\uparrow$0.82} & \up{19.92}{3.52} & \up{27.95}{7.48} & \avg{$\uparrow$5.50} \\
      & Few-Shot (BM25)       & \up{36.71}{0.84} & \up{53.84}{1.57} & \up{26.72}{2.13} & \avg{$\uparrow$1.51} & \up{20.70}{5.08} & \up{27.56}{4.72} & \avg{$\uparrow$4.90} \\
    \midrule

    \multirow{5.5}{*}{\rotatebox{90}{\textbf{Qwen2-7B}}} 
      & Zero-Shot & 32.49 & 50.13 & 26.25 & - & 21.09 & 35.83 & - \\
      \cdashline{2-9} \addlinespace[2pt]
      & Few-Shot (Rand. Std.) & \up{38.54}{1.69} & \up{56.39}{1.99} & \up{29.94}{2.26} & \avg{$\uparrow$1.98} & \up{41.02}{0.78} & \up{55.51}{1.97} & \avg{$\uparrow$1.38} \\
      & Few-Shot (Rand. Ext.) & \up{42.33}{0.99} & \up{59.43}{1.69} & \up{31.62}{1.61} & \avg{$\uparrow$1.43} & \up{36.33}{2.73} & \up{53.15}{1.57} & \avg{$\uparrow$2.15} \\
      & Few-Shot (KATE)       & \up{43.32}{0.28} & \up{60.56}{2.18} & \up{32.69}{1.58} & \avg{$\uparrow$1.35} & \up{43.36}{6.64} & \up{49.21}{4.33} & \avg{$\uparrow$5.49} \\
      & Few-Shot (BM25)       & \up{42.48}{0.42} & \up{59.34}{1.39} & \up{31.30}{1.48} & \avg{$\uparrow$1.10} & \up{48.83}{0.78} & \up{53.54}{0.40} & \avg{$\uparrow$0.59} \\
    \midrule
    
    \multirow{5.5}{*}{\rotatebox{90}{\textbf{Yi-6B}}} 
      & Zero-Shot & 25.46 & 44.64 & 21.50 & - & 3.52 & 12.99 & - \\
      \cdashline{2-9} \addlinespace[2pt]
      & Few-Shot (Rand. Std.) & \up{34.18}{1.26} & \up{52.01}{2.20} & \up{26.67}{1.32} & \avg{$\uparrow$1.59} & \up{\phantom{0}5.47}{0.00} & \down{32.28}{0.78} & \ravg{$\downarrow$0.39} \\
      & Few-Shot (Rand. Ext.) & \up{38.82}{2.81} & \up{55.14}{2.74} & \up{28.51}{2.36} & \avg{$\uparrow$2.64} & \up{\phantom{0}5.08}{0.17} & \up{32.28}{4.73} & \avg{$\uparrow$2.45} \\
      & Few-Shot (KATE)       & \up{38.68}{3.09} & \up{55.66}{2.57} & \up{28.68}{2.71} & \avg{$\uparrow$2.79} & \up{17.19}{1.56} & \up{27.56}{5.51} & \avg{$\uparrow$3.54} \\
      & Few-Shot (BM25)       & \up{38.68}{2.53} & \up{54.43}{2.86} & \up{27.85}{2.51} & \avg{$\uparrow$2.63} & \up{17.58}{3.12} & \up{31.89}{0.79} & \avg{$\uparrow$1.96} \\
    \midrule

    \multirow{5.5}{*}{\rotatebox{90}{\textbf{Llama-2-7B}}} 
      & Zero-Shot & 26.58 & 44.94 & 21.76 & - & 0.00 & 7.48 & - \\
      \cdashline{2-9} \addlinespace[2pt]
      & Few-Shot (Rand. Std.) & \up{36.01}{1.54} & \up{54.97}{2.33} & \up{27.92}{2.28} & \avg{$\uparrow$2.05} & \up{\phantom{0}1.95}{1.57} & \up{19.69}{1.57} & \avg{$\uparrow$1.57} \\
      & Few-Shot (Rand. Ext.) & \up{39.66}{2.25} & \up{57.36}{1.76} & \up{29.32}{2.57} & \avg{$\uparrow$2.19} & \up{\phantom{0}2.73}{2.35} & \up{19.69}{0.78} & \avg{$\uparrow$1.57} \\
      & Few-Shot (KATE)       & \up{39.94}{1.97} & \up{57.06}{2.25} & \up{29.16}{2.92} & \avg{$\uparrow$2.38} & \up{\phantom{0}9.77}{6.64} & \up{16.93}{1.18} & \avg{$\uparrow$3.91} \\
      & Few-Shot (BM25)       & \up{39.38}{2.53} & \up{57.12}{2.65} & \up{28.66}{2.66} & \avg{$\uparrow$2.61} & \up{12.50}{3.91} & \up{17.72}{1.57} & \avg{$\uparrow$2.74} \\
    \midrule

    \multirow{5.5}{*}{\rotatebox{90}{\textbf{Llama-3-8B}}} 
      & Zero-Shot & 37.83 & 58.06 & 30.69 & - & 0.00 & 28.74 & - \\
      \cdashline{2-9} \addlinespace[2pt]
      & Few-Shot (Rand. Std.) & \up{44.87}{2.81} & \up{65.76}{1.32} & \up{36.87}{1.76} & \avg{$\uparrow$1.96} & \up{18.75}{2.73} & \up{41.34}{2.36} & \avg{$\uparrow$2.55} \\
      & Few-Shot (Rand. Ext.) & \up{49.93}{0.84} & \up{68.25}{1.47} & \up{39.86}{1.18} & \avg{$\uparrow$1.16} & \up{21.88}{0.39} & \up{42.13}{3.93} & \avg{$\uparrow$2.16} \\
      & Few-Shot (KATE)       & \up{48.10}{1.83} & \up{67.53}{1.32} & \up{38.89}{1.88} & \avg{$\uparrow$1.68} & \down{37.11}{0.78} & \up{46.06}{6.69} & \avg{$\uparrow$2.96} \\
      & Few-Shot (BM25)       & \up{48.52}{1.13} & \up{67.10}{0.59} & \up{38.61}{0.75} & \avg{$\uparrow$0.82} & \up{33.20}{1.57} & \up{44.09}{0.79} & \avg{$\uparrow$1.18} \\
    \midrule

    \multirow{5.5}{*}{\rotatebox{90}{\textbf{Gemma-2-9B}}} 
      & Zero-Shot & 27.00 & 45.06 & 22.08 & - & 5.47 & 6.69 & - \\
      \cdashline{2-9} \addlinespace[2pt]
      & Few-Shot (Rand. Std.) & \up{39.24}{1.69} & \up{56.11}{1.93} & \up{28.55}{2.02} & \avg{$\uparrow$1.88} & \up{21.09}{5.47} & \up{44.88}{5.51} & \avg{$\uparrow$5.49} \\
      & Few-Shot (Rand. Ext.) & \up{45.43}{0.84} & \up{63.04}{0.69} & \up{33.64}{0.81} & \avg{$\uparrow$0.78} & \up{26.56}{1.17} & \up{46.85}{1.58} & \avg{$\uparrow$1.38} \\
      & Few-Shot (KATE)       & \up{46.55}{1.55} & \up{63.28}{0.63} & \up{34.17}{1.00} & \avg{$\uparrow$1.06} & \up{35.94}{1.95} & \up{47.24}{2.76} & \avg{$\uparrow$2.36} \\
      & Few-Shot (BM25)       & \up{43.32}{1.12} & \up{60.93}{1.25} & \up{32.00}{1.08} & \avg{$\uparrow$1.15} & \up{37.89}{4.30} & \up{46.85}{0.39} & \avg{$\uparrow$2.35} \\

    \bottomrule
  \end{tabular}
  \caption{\textbf{Main Results.} Performance on TruthfulQA, Math-500, and AQUA-RAT across seven LLMs. Values are formatted as: Baseline (\cbox{$\uparrow$\textbf{Gain}} or \rbox{$\downarrow$Loss} compared to baseline with our \textsc{DeCoVec} during decoding).}
  \label{tab:main_results}
\end{table*}

The refined task-enhanced decoding logit $\tilde{\mathbf{z}}^{t} = \{{\tilde{\mathbf{z}}^{t}_{[1]}, \tilde{\mathbf{z}}^{t}_{[2]}, ..., \tilde{\mathbf{z}}^{t}_{[{|\mathcal{V}|}]}}\}$ is used to replace the original logits for generating the token in step $t+1$ from vocabulary $\mathcal{V} = \{v_{[1]}, v_{[2]}, ..., v_{[{|\mathcal{V}|}]}\}$:
\begin{align}
    {\rm idx} = {\rm argmax}_k\ &\tilde{\mathbf{z}}^{t}_{[k]}, k=1,2,...,|\mathcal{V}|,\\
    \tilde{y}^{t+1} &= v_{[{\rm idx}]} \in \mathcal{V}.
\end{align}
This steering effectively aligns generation with distilled task semantics, yielding the correct prediction shown in Figure~\ref{fig:decovec_pipeline}(c).

\section{Experiments}

\subsection{Settings}

\noindent\textbf{Datasets.} 
We use \textbf{TruthfulQA} \cite{lin-etal-2022-truthfulqa} to evaluate the model’s propensity to generate truthful answers. We also include two reasoning benchmarks: \textbf{MATH-500} \cite{lightman_let_2024_math500}, a representative subset of the challenging MATH dataset, and \textbf{AQUA-RAT} \cite{ling-etal-2017-program-aquarat}, a dataset consisting of algebra word problems with rationales.

\noindent\textbf{Contexts for ICL.} 
To simulate a data-efficient scenario, we restrict the size of the available demonstrations during ICL. For each dataset, we randomly sample only \textbf{100 examples} to form a \textbf{candidate pool} used for both exemplar retrieval (for baselines and ours) and hyperparameter selection. 

For TruthfulQA, we use 100 candidates and the remaining 690 samples for evaluation. For MATH-500, we use 100 candidates and 256 samples from the remaining 400 for evaluation. For AQUA-RAT, we use 100 candidates from validation set and 254 samples from test set for evaluation.

\noindent\textbf{Models.}
We select advanced open-source LLMs ranging from 0.5B to 9B parameters to thoroughly verify scalability and generalization, including \textbf{Qwen2} family (0.5B, 1.5B, 7B; \citealp{yang_qwen2_2024}), \textbf{Yi-6B} \cite{ai_yi_2025}, \textbf{Llama-2-7B} \cite{touvron_llama_2023-llama2}, \textbf{Llama-3-8B} \cite{grattafiori2024llama}, and \textbf{Gemma-2-9B} \cite{team_gemma_2024}.

\noindent\textbf{Demonstrations.} We apply three demonstration selection strategies that do not require additional training. \textbf{Random Selection} samples in-context examples uniformly from the candidate pool for each query. \textbf{KATE} \cite{liu2021KATE} retrieves semantically similar exemplars based on the cosine similarity of demonstration embeddings, utilizing the \texttt{all-MiniLM-L6-v2} model~\cite{wang_minilmv2_2021}. \textbf{BM25} \cite{robertson2009BM25} performs sparse retrieval to select exemplars based on lexical overlap and keyword matching.

In the few-shot setting, we set \textit{standard} (std.) and \textit{extended} (ext.) number of demonstration $k$. For TruthfulQA, we set $k=20$ for standard and $k=50$ for extended settings. For MATH-500 and AQUA-RAT that require long reasoning chains, we set $k=2$ for standard and $k=10$ for extended settings due to context window constraints.

\noindent\textbf{Implementation Details.}
We use greedy decoding for all tasks. Steering intensity is fixed at $\lambda = 1.0$ for TruthfulQA. For Math-500 and AQUA-RAT, we select $\lambda \in \{0.2, 0.4, \dots, 1.0\}$ on the candidate pool, which serves as a validation set, and apply the chosen $\lambda$ at test time. All experiments run on NVIDIA RTX 5090$\times$8 GPUs.
More details are shown in Appendix~\ref{sec:add_details}.

\subsection{Main Results}
\label{sec:main_results}

We present the comprehensive performance comparison in Table~\ref{tab:main_results}. 
Consistent with established findings, few-shot baselines significantly outperform zero-shot settings across all models (e.g., Qwen2-0.5B gains over 7\% on TruthfulQA). This confirms that the provided contexts successfully activate latent task capabilities, serving as the necessary foundation for our subsequent vector extraction.
We highlight further in-depth observations below:

\paragraph{\textsc{DeCoVec} demonstrates universal compatibility and scale-agnostic effectiveness.}
Our proposed \textsc{DeCoVec} exhibits robust adaptability across diverse settings. By steering the decoding process with the extracted task vector, \textsc{DeCoVec} achieves performance gains in almost all evaluated scenarios, with the average improvement ($\Delta$) ranging from 0.59 to 5.50. 
While we observe a slight performance dip for Yi-6B on reasoning tasks, \textsc{DeCoVec} delivers substantial gains in the majority of cases, particularly achieving a remarkable 5.50 average increase on Qwen2-1.5B.
Notably, this benefit is not strictly proportional to model size; small models benefit just as significantly as larger ones.
Calculating the average $\Delta$ across all four baselines on TruthfulQA, the tiny Qwen2-0.5B achieves an average gain of 1.58, which is comparable to (and even slightly exceeds) the 1.47 gain of the much larger Qwen2-7B. 
A similar trend holds on reasoning tasks (2.51 vs. 2.40).
This consistency suggests that \textsc{DeCoVec} effectively extracts and amplifies the task representations inherent to the model's specific capacity, regardless of the parameter count.

\paragraph{\textsc{DeCoVec} performance is sensitive to retrieval but decoupled from baseline metrics.}
We observe that the optimal demonstration selection strategy varies by domain: TruthfulQA favors semantic retrieval (KATE), while reasoning datasets benefit more from lexical overlap (BM25). 
Crucially, we find that the improvement provided by \textsc{DeCoVec} is \textbf{not strictly tied to the performance gap between few-shot and zero-shot, nor to the absolute few-shot accuracy}. Instead, the gain reflects a complex interplay between the extraction of task semantics and the inherent difficulty of the remaining errors.
On one hand, richer semantic contexts facilitate better and more stable vector extraction; for instance, on Yi-6B (TruthfulQA), while BM25 and KATE yield the exact same baseline (38.68), the semantically richer KATE context enables a larger \textsc{DeCoVec} boost (3.09) compared to BM25 (2.53).
On the other hand, a lower initial baseline does not preclude superior final performance. On Yi-6B (AQUA-RAT), the KATE baseline (27.56) significantly lags behind BM25 (31.89). However, \textsc{DeCoVec} extracts a more potent direction from the semantically-rich KATE context, not only providing a massive gain but ultimately \textbf{propelling the final accuracy (33.07) above that of the BM25-steered model (32.68)}.

\paragraph{\textsc{DeCoVec} amplifies high-quality retrieval and mitigates noise in extended contexts.}
For TruthfulQA, increasing the number of shots generally improves stability. However, retrieval-based methods (KATE) often achieve superior performance with fewer shots ($k=20$) compared to random extended shots ($k=50$). \textsc{DeCoVec} further amplifies this advantage, yielding the best results when combined with high-quality retrieval.
For Math-500 and AQUA-RAT, the trend is less clear. Simply adding more random shots (Rand. Std. $\to$ Rand. Ext.) does not guarantee improvement and often degrades performance (e.g., Qwen2-7B drops from 41.02 to 36.33 on Math-500), likely due to noise introduced in the extended context. In these noise-sensitive scenarios, \textsc{DeCoVec} shows strong robustness, effectively recovering performance (e.g., lifting Qwen2-7B back to 39.06) by distilling essential task direction from noisy input.

\definecolor{graybg}{gray}{0.9} 
\definecolor{azure}{rgb}{0.94, 1.0, 1.0}

\begin{table}[t!]
  \centering
  \small
  \setlength{\tabcolsep}{5pt} 
  \renewcommand{\arraystretch}{1.2}
  
  \begin{tabular}{l ccc}
    \toprule
    \multirow{2}{*}{\textbf{Methods}} & \textbf{TruthfulQA} & \textbf{Math-500} & \textbf{AQUA-RAT} \\
     & {Avg. MC} & {Acc.} & {Acc.} \\
    \midrule
    
    \addlinespace[1pt]
    \rowcolor{graybg}
    \multicolumn{4}{c}{\textit{Sequential Order}} \\
    KATE & 45.52\phantom{$_{\uparrow 0.00}$} & 43.36\phantom{$_{\uparrow 0.00}$} & 49.21\phantom{$_{\uparrow 0.00}$} \\
    \rowcolor{azure}
    \textit{w/} \textsc{DeCoVec} & \textbf{46.87}$_{\uparrow 1.35}$ & \textbf{50.00}$_{\uparrow 6.64}$ & \textbf{53.54}$_{\uparrow 4.33}$ \\
    BM25 & 44.37\phantom{$_{\uparrow 0.00}$} & 48.83\phantom{$_{\uparrow 0.00}$} & 53.54\phantom{$_{\uparrow 0.00}$} \\
    \rowcolor{azure}
    \textit{w/} \textsc{DeCoVec} & \textbf{45.47}$_{\uparrow 1.10}$ & \textbf{49.61}$_{\uparrow 0.78}$ & \textbf{53.94}$_{\uparrow 0.40}$ \\
    
    \midrule
    \addlinespace[1pt]
    \rowcolor{graybg}
    \multicolumn{4}{c}{\textit{Reverse Order}} \\
    KATE & 45.38\phantom{$_{\uparrow 0.00}$} & 49.22\phantom{$_{\uparrow 0.00}$} & 48.82\phantom{$_{\uparrow 0.00}$} \\
    \rowcolor{azure}
    \textit{w/} \textsc{DeCoVec} & \textbf{46.69}$_{\uparrow 1.31}$ & 48.83$_{\downarrow 0.39}$ & \textbf{52.36}$_{\uparrow 3.54}$ \\
    BM25 & 44.21\phantom{$_{\uparrow 0.00}$} & 50.78\phantom{$_{\uparrow 0.00}$} & 54.72\phantom{$_{\uparrow 0.00}$} \\
    \rowcolor{azure}
    \textit{w/} \textsc{DeCoVec} & \textbf{45.70}$_{\uparrow 1.49}$ & \textbf{52.73}$_{\uparrow 1.95}$ & \textbf{55.12}$_{\uparrow 0.40}$ \\
    
    \midrule
    \addlinespace[1pt]
    \rowcolor{graybg}
    \multicolumn{4}{c}{\textit{Random Order}} \\
    KATE & 45.48\phantom{$_{\uparrow 0.00}$} & 45.31\phantom{$_{\uparrow 0.00}$} & 49.08\phantom{$_{\uparrow 0.00}$} \\
    \rowcolor{azure}
    \textit{w/} \textsc{DeCoVec} & \textbf{46.63}$_{\uparrow 1.15}$ & \textbf{49.61}$_{\uparrow 4.30}$ & \textbf{53.15}$_{\uparrow 4.07}$ \\
    BM25 & 44.73\phantom{$_{\uparrow 0.00}$} & 49.48\phantom{$_{\uparrow 0.00}$} & 53.93\phantom{$_{\uparrow 0.00}$} \\
    \rowcolor{azure}
    \textit{w/} \textsc{DeCoVec} & \textbf{45.56}$_{\uparrow 0.83}$ & \textbf{50.65}$_{\uparrow 1.17}$ & \textbf{54.33}$_{\uparrow 0.40}$ \\
    
    \bottomrule
  \end{tabular}
  \caption{\textbf{Results with different order of examples.} \newline \textsc{DeCoVec} provides consistent gains across most ordering strategies and datasets.} 
  \label{tab:ablation_order_combined}
\end{table}
\section{Analysis}
\label{sec:analysis}

We discuss key factors for building our \textsc{DeCoVec} as well as error distribution and token efficiency.

\subsection{Impact of Demonstrations in ICL}

\paragraph{Sensitivity to the Order of Examples.}
ICL performance often suffers from recency bias due to demonstration ordering~\cite{pmlr-v139-zhao21c}.  
We evaluate \textsc{DeCoVec} on Qwen2-7B with \textit{sequential}, \textit{reverse}, and \textit{random} orders (Table~\ref{tab:ablation_order_combined}).  
While baselines fluctuate (e.g., KATE ranges from 43.36\% to 49.22\% on Math-500), \textsc{DeCoVec} consistently improves performance—by up to 6.64\% in the worst case—showing robustness to input arrangement via invariant task semantics distillation.

\begin{figure}[t]
    \centering
    \includegraphics[width=\columnwidth]{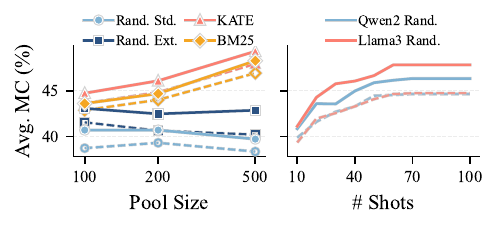} 
    \caption{
        \textbf{Impact of demonstration settings on TruthfulQA.}
        Solid lines: \textsc{DeCoVec}; dashed lines: few-shot baselines.
        Left: Sensitivity to candidate pool size.
        Right: Scalability with respect to the number of shots.
    }
    \label{fig:demo_analysis}
\end{figure}

\paragraph{Impact of Candidate Pool Size.}

The candidate pool size directly impacts context quality, as shown in Figure~\ref{fig:demo_analysis} (left). Applying demonstration selection strategies KATE and BM25 scales positively with pool size, benefiting from a richer selection of relevant exemplars. Conversely, random baselines degrade due to low relevance, as larger pools increase the risk of sampling irrelevant noise. Overall, \textsc{DeCoVec} yields consistent gains across all settings, demonstrating its robustness in extracting valid task directions even from noisy contexts.
This decoupling from strict retrieval quality is particularly valuable in resource-constrained scenarios where building a high-precision retrieval index is not feasible. \textsc{DeCoVec} acts as a denoising filter, amplifying the useful task signal even when the retrieved context is suboptimal.

\begin{figure*}[t!]
    \centering
    \includegraphics[width=\textwidth]{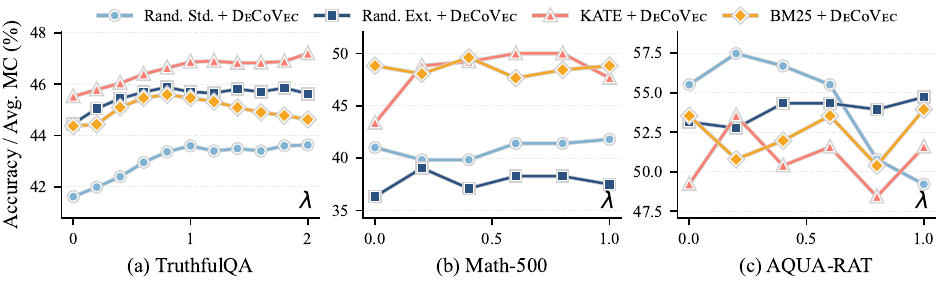} 
    \caption{\textbf{Sensitivity analysis on hyperparameter $\lambda$}. 
    (a) TruthfulQA remains robust to large $\lambda$ ($\geq 1.0$); (b,c) Math-500 and AQUA-RAT performance degrades if $\lambda$ is poorly calibrated. Appendix~\ref{sec:add_details} lists the calibrated values.}
    \label{fig:main_ablation}
\end{figure*}

\paragraph{Impact of Shot Quantity.}

We further investigate the impact of the number of demonstrations $k$. As shown in Figure~\ref{fig:demo_analysis} (right), increasing $k$ from 10 to 60 generally improves the baseline performance (dashlines) before plateauing. Overall, \textsc{DeCoVec} (solid lines) maintains a consistent and significant performance lead across all shot settings for both Qwen-2-7B and Llama-3-8B.


\subsection{Impact of $\lambda$ }
\label{subsec:lambda}

The hyperparameter $\lambda$ in Eq.~\ref{eq:decoding_steering} governs the intensity of the task vector injection. As illustrated in Figure~\ref{fig:main_ablation}, the distinct behaviors of different tasks validate our experimental configurations.

For {TruthfulQA} (Figure~\ref{fig:main_ablation}a), performance improves rapidly and remains stable even at higher values ($\lambda \ge 1.0$). This indicates that the ``truthfulness'' direction is robust and broad, allowing for strong steering without degradation. In contrast, {Mathematical Reasoning Tasks} (Math-500 in Figure~\ref{fig:main_ablation}b and AQUA-RAT in Figure~\ref{fig:main_ablation}c) exhibit marked sensitivity to $\lambda$. Unlike open-ended generation, reasoning relies on rigorous logic chains and precise numerical calculations, which are easily disrupted by excessive intervention. 

\begin{table}[t!]
    \centering
    \small
    \setlength{\tabcolsep}{2.8pt} 
    \begin{tabular}{lccccc}
        \toprule
        \textbf{Methods} & \textbf{LR} & \textbf{DG} & \textbf{CA} & \textbf{CM} & \textbf{Total Err.} \\
        \midrule
        KATE & 35.4\% & 9.4\% & \textbf{6.5\%} & \textbf{5.3\%} & 56.6\% \\
        \cc{\textit{w}/ \textsc{DeCoVec}} & \cc{\textbf{27.0\%}} & \cc{\textbf{7.3\%}} & \cc{8.8\%} & \cc{6.9\%} & \cc{\textbf{50.0\%}} \\
        \textit{$\Delta$} & \textit{\textcolor{green!60!black}{$\downarrow$8.4\%}} & \textit{\textcolor{green!60!black}{$\downarrow$2.1\%}} & \textit{\textcolor{red}{$\uparrow$2.3\%}} & \textit{\textcolor{red}{$\uparrow$1.6\%}} & \textit{\textcolor{green!60!black}{$\downarrow$6.6\%}} \\
        \bottomrule
    \end{tabular}
    \caption{\textbf{Distribution on different error types.} \newline LR: logical and reasoning flaws. DG: degeneration. \newline CA: calculation. CM: concept and formula misuse. }

    \label{tab:error_analysis}
\end{table}

\subsection{Error Distribution}
\label{subsec:error_analysis}

To investigate whether \textsc{DeCoVec} effectively captures and encodes task-level information as hypothesized, we conducted a fine-grained error analysis on the Qwen2-7B model outputs using Math-500 dataset. Following \citet{lightman_let_2024_math500}, we employed an LLM-based evaluator to categorize incorrect solutions into four types: logical/reasoning flaws (LR), degeneration (DG), calculation errors (CA), and concept/formula misuse (CM).

As shown in Table~\ref{tab:error_analysis}, applying \textsc{DeCoVec} results in a distinct shift in the error distribution: logical flaws and degeneration decrease significantly, whereas calculation errors and concept misuse rise slightly. This trade-off provides compelling evidence for our hypothesis: \textsc{DeCoVec} primarily extracts task-level semantics, representing the high-level understanding of the task and problem-solving strategies, rather than fine-grained numerical control. In other words, the vector steers the model to ``think'' more logically and coherently, even if this semantic steering comes at a minor cost to the precision of local calculations. 

\subsection{Inference Cost}
\label{subsec:inference_cost}
We compare the average number of output tokens generated by the baseline and our approach on Math-500 using Qwen2-7B, as shown in Table~\ref{tab:token_cost}. We find that traditional in-context learning methods tend to produce irrelevant and repetitive content, whereas our method can reduce the generated tokens by 10$\sim$40 across various settings. Overall, although our decoding-based method introduces additional lightweight computational overhead, it can alleviate the extra inference cost by substantially reducing the output length. Detailed end-to-end latency measurements showing an empirical overhead of $1.6\times$ to $1.7\times$ are provided in Appendix~\ref{app:inference_overhead}.

\begin{table}[t!]
    \centering
    \small
\setlength{\tabcolsep}{2.8pt} 
    \begin{tabular}{lcccc}
        \toprule
        \textbf{Methods} & \textbf{Rand. Std.} & \textbf{Rand. Ext.}& \textbf{KATE}& \textbf{BM25} \\
        \midrule
        Few-Shot ICL & 354.8& 344.2& 326.1& 324.6  \\
        \cc{\textit{w}/ \textsc{DeCoVec} & \cc{\textbf{314.5}}}& \cc{\textbf{333.6}}& \cc{\textbf{312.3}} & \cc{\textbf{300.1}} \\
        \textit{$\Delta$} & \textit{\textcolor{green!60!black}{$\downarrow$40.3}}&\textit{\textcolor{green!60!black}{$\downarrow$10.6}}&\textit{\textcolor{green!60!black}{$\downarrow$13.8}}& \textit{\textcolor{green!60!black}{$\downarrow$24.5}}\\
        \bottomrule
    \end{tabular}
    \caption{\textbf{Comparison of average output tokens.} \textsc{DeCoVec} reduces the generation length across all settings.}
    \label{tab:token_cost}
\end{table}


\section{Conclusion}
We introduced \textsc{DeCoVec}, a novel framework for constructing task vectors within the decoding space.
By contrasting the logit distributions of few-shot and zero-shot contexts, \textsc{DeCoVec} distills task-specific capabilities into a lightweight vector that steers model behavior during decoding.
As a non-invasive and training-free approach, it avoids the computational overhead of parameter tuning and the complexity of internal activation manipulation.
Our extensive experiments demonstrate that \textsc{DeCoVec} effectively enhances performance across diverse tasks and models, offering a flexible and efficient method for steering large language models.

\section*{Limitations}

Despite the effectiveness of \textsc{DeCoVec}, three limitations remain for future exploration. 

First, the scaling factor $\lambda$ is currently tuned at the dataset level, which overlooks the varying sensitivity of different input instances to steering. Future work aims to develop adaptive mechanisms to dynamically calibrate the steering intensity based on instance-level uncertainty or prediction entropy, ensuring an optimal trade-off between task enhancement and generation quality.

Second, the applicability of \textsc{DeCoVec} to reasoning-specialized models with long thought processes remains to be verified. These models rely heavily on the coherence of internal states across multiple reasoning steps, so it is crucial to investigate whether decoding-space interventions preserve chain-of-thought stability or require tailored adjustment strategies for sequential reasoning.

Third, the compositional arithmetic properties of decoding-space vectors remain underexplored compared to their weight-space counterparts. Future work could investigate merging distinct \textsc{DeCoVec} vectors, such as combining reasoning and safety attributes, to simultaneously steer models toward multiple downstream objectives.

\section*{Acknowledgments}
This work was supported in parts by Guangdong Basic and Applied Basic Research Foundation (2026A1515011358), Shenzhen Natural Science Foundation (JCYJ20250604181610014), and Intelligent Computing Center of Shenzhen University.

\bibliography{custom}

@inproceedings{
wang2023selfconsistency,
title={Self-Consistency Improves Chain of Thought Reasoning in Language Models},
author={Xuezhi Wang and Jason Wei and Dale Schuurmans and Quoc V Le and Ed H. Chi and Sharan Narang and Aakanksha Chowdhery and Denny Zhou},
booktitle={The Eleventh International Conference on Learning Representations },
year={2023},
url={https://openreview.net/forum?id=1PL1NIMMrw}
}

@article{ai_yi_2025,
  title={Yi: Open foundation models by 01. ai},
  url = {http://arxiv.org/abs/2403.04652},
  author={Young, Alex and Chen, Bei and Li, Chao and Huang, Chengen and Zhang, Ge and Zhang, Guanwei and Wang, Guoyin and Li, Heng and Zhu, Jiangcheng and Chen, Jianqun and others},
  journal={arXiv preprint arXiv:2403.04652},
  year={2024}
}

@article{touvron_llama_2023-llama2,
  title={Llama 2: Open foundation and fine-tuned chat models},
  url = {http://arxiv.org/abs/2307.09288},
  doi = {10.48550/arXiv.2307.09288},
  author={Touvron, Hugo and Martin, Louis and Stone, Kevin and Albert, Peter and Almahairi, Amjad and Babaei, Yasmine and Bashlykov, Nikolay and Batra, Soumya and Bhargava, Prajjwal and Bhosale, Shruti and others},
  journal={arXiv preprint arXiv:2307.09288},
  year={2023}
}

@article{team_gemma_2024,
  title={Gemma 2: Improving open language models at a practical size},
	url = {http://arxiv.org/abs/2408.00118},
	doi = {10.48550/arXiv.2408.00118},
  author={{Gemma Team}},
  journal={arXiv preprint arXiv:2408.00118},
  year={2024}
}

@article{yang_qwen2_2024,
  title={Qwen2 technical report},
	url = {http://arxiv.org/abs/2407.10671},
	doi = {10.48550/arXiv.2407.10671},
  author={{Qwen Team}},
  journal={arXiv preprint arXiv:2407.10671},
  volume={2},
  number={3},
  year={2024}
}

@inproceedings{lin-etal-2022-truthfulqa,
    title = "{T}ruthful{QA}: Measuring How Models Mimic Human Falsehoods",
    author = "Lin, Stephanie  and
      Hilton, Jacob  and
      Evans, Owain",
    editor = "Muresan, Smaranda  and
      Nakov, Preslav  and
      Villavicencio, Aline",
    booktitle = "Proceedings of the 60th Annual Meeting of the Association for Computational Linguistics (Volume 1: Long Papers)",
    month = may,
    year = "2022",
    address = "Dublin, Ireland",
    publisher = "Association for Computational Linguistics",
    url = "https://aclanthology.org/2022.acl-long.229/",
    doi = "10.18653/v1/2022.acl-long.229",
    pages = "3214--3252"
}

@article{grattafiori2024llama,
  title={The {L}lama 3 herd of models},
  url={https://arxiv.org/abs/2407.21783},
  doi={10.48550/arXiv.2407.21783},
  author={Grattafiori, Aaron and Dubey, Abhimanyu and Jauhri, Abhinav and Pandey, Abhinav and Kadian, Abhishek and Al-Dahle, Ahmad and Letman, Aiesha and Mathur, Akhil and Schelten, Alan and Vaughan, Alex and others},
  journal={arXiv preprint arXiv:2407.21783},
  year={2024}
}

@inproceedings{lightman_let_2024_math500,
  title={Let's verify step by step},
  url = {https://proceedings.iclr.cc/paper_files/paper/2024/file/aca97732e30bcf1303bc22ac3924fd16-Paper-Conference.pdf},
  author={Lightman, Hunter and Kosaraju, Vineet and Burda, Yuri and Edwards, Harrison and Baker, Bowen and Lee, Teddy and Leike, Jan and Schulman, John and Sutskever, Ilya and Cobbe, Karl},
  booktitle={The Twelfth International Conference on Learning Representations},
  year={2023}
}

@inproceedings{ling-etal-2017-program-aquarat,
    title = "Program Induction by Rationale Generation: Learning to Solve and Explain Algebraic Word Problems",
    author = "Ling, Wang  and
      Yogatama, Dani  and
      Dyer, Chris  and
      Blunsom, Phil",
    editor = "Barzilay, Regina  and
      Kan, Min-Yen",
    booktitle = "Proceedings of the 55th Annual Meeting of the Association for Computational Linguistics (Volume 1: Long Papers)",
    month = jul,
    year = "2017",
    address = "Vancouver, Canada",
    publisher = "Association for Computational Linguistics",
    url = "https://aclanthology.org/P17-1015/",
    doi = "10.18653/v1/P17-1015",
    pages = "158--167"
}

@inproceedings{wang_minilmv2_2021,
    title = "{M}ini{LM}v2: Multi-Head Self-Attention Relation Distillation for Compressing Pretrained Transformers",
    author = "Wang, Wenhui  and
      Bao, Hangbo  and
      Huang, Shaohan  and
      Dong, Li  and
      Wei, Furu",
    editor = "Zong, Chengqing  and
      Xia, Fei  and
      Li, Wenjie  and
      Navigli, Roberto",
    booktitle = "Findings of the Association for Computational Linguistics: ACL-IJCNLP 2021",
    month = aug,
    year = "2021",
    address = "Online",
    publisher = "Association for Computational Linguistics",
    url = "https://aclanthology.org/2021.findings-acl.188/",
    doi = "10.18653/v1/2021.findings-acl.188",
    pages = "2140--2151"
}

@book{robertson2009BM25,
author = {Robertson, Stephen and Zaragoza, Hugo},
title = {The Probabilistic Relevance Framework},
url = {https://dl.acm.org/doi/abs/10.5555/1823431},
year = {2009},
isbn = {1601983085},
publisher = {Now Publishers Inc.},
address = {Hanover, MA, USA}
}

@inproceedings{liu2021KATE,
 author = {Zhang, Yuanhan and Zhou, Kaiyang and Liu, Ziwei},
 booktitle = {Advances in Neural Information Processing Systems},
 editor = {A. Oh and T. Naumann and A. Globerson and K. Saenko and M. Hardt and S. Levine},
 pages = {17773--17794},
 publisher = {Curran Associates, Inc.},
 title = {What Makes Good Examples for Visual In-Context Learning?},
 url = {https://proceedings.neurips.cc/paper_files/paper/2023/file/398ae57ed4fda79d0781c65c926d667b-Paper-Conference.pdf},
 volume = {36},
 year = {2023}
}

@article{yang2025task,
  title={Task vectors in in-context learning: Emergence, formation, and benefit},
  url= {https://arxiv.org/abs/2501.09240},
  author={Yang, Liu and Lin, Ziqian and Lee, Kangwook and Papailiopoulos, Dimitris and Nowak, Robert},
  journal={arXiv preprint arXiv:2501.09240},
  year={2025}
}

@inproceedings{hojel2024finding_visualtaskvectors,
author = {Hojel, Alberto and Bai, Yutong and Darrell, Trevor and Globerson, Amir and Bar, Amir},
title = {Finding Visual Task Vectors},
year = {2024},
isbn = {978-3-031-72774-0},
publisher = {Springer-Verlag},
address = {Berlin, Heidelberg},
url = {https://doi.org/10.1007/978-3-031-72775-7_15},
doi = {10.1007/978-3-031-72775-7_15},
booktitle = {Computer Vision – ECCV 2024: 18th European Conference, Milan, Italy, September  29–October 4, 2024, Proceedings,  Part XLIII},
pages = {257–273},
numpages = {17},
location = {Milan, Italy}
}

@article{fierro2025steering_constructvector,
  title={Steering Language Models with Weight Arithmetic},
  url={https://arxiv.org/abs/2511.05408},
  author={Fierro, Constanza and Roger, Fabien},
  journal={arXiv preprint arXiv:2511.05408},
  year={2025}
}

@inproceedings{kalyan2024emotion,
  title     = {{Emotion Arithmetic: Emotional Speech Synthesis via Weight Space Interpolation}},
  author    = {Pavan Kalyan and Preeti Rao and Preethi Jyothi and Pushpak Bhattacharyya},
  year      = {2024},
url = {https://www.ee.iitb.ac.in/course/~daplab/publications/2024/kalyan24_interspeech.pdf},
  booktitle = {{Interspeech 2024}},
  pages     = {1805--1809},
  doi       = {10.21437/Interspeech.2024-2311},
  issn      = {2958-1796},
}

@inproceedings{huang-etal-2024-chat-chatvector,
    title = "Chat Vector: A Simple Approach to Equip {LLM}s with Instruction Following and Model Alignment in New Languages",
    author = "Huang, Shih-Cheng  and
      Li, Pin-Zu  and
      Hsu, Yu-chi  and
      Chen, Kuang-Ming  and
      Lin, Yu Tung  and
      Hsiao, Shih-Kai  and
      Tsai, Richard  and
      Lee, Hung-yi",
    editor = "Ku, Lun-Wei  and
      Martins, Andre  and
      Srikumar, Vivek",
    booktitle = "Proceedings of the 62nd Annual Meeting of the Association for Computational Linguistics (Volume 1: Long Papers)",
    month = aug,
    year = "2024",
    address = "Bangkok, Thailand",
    publisher = "Association for Computational Linguistics",
    url = "https://aclanthology.org/2024.acl-long.590/",
    doi = "10.18653/v1/2024.acl-long.590",
    pages = "10943--10959"
}

@inproceedings{
hu_lora_2021,
title={Lo{RA}: Low-Rank Adaptation of Large Language Models},
author={Edward J Hu and yelong shen and Phillip Wallis and Zeyuan Allen-Zhu and Yuanzhi Li and Shean Wang and Lu Wang and Weizhu Chen},
booktitle={International Conference on Learning Representations},
year={2022},
url={https://openreview.net/forum?id=nZeVKeeFYf9}
}

@inproceedings{
hu2025distributionaligned,
title={Distribution-Aligned Decoding for Efficient {LLM} Task Adaptation},
author={Senkang Hu and Xudong Han and Jinqi Jiang and Yihang Tao and Zihan Fang and Yong Dai and Sam Kwong and Yuguang Fang},
booktitle={The Thirty-ninth Annual Conference on Neural Information Processing Systems},
year={2025},
url={https://openreview.net/forum?id=IVWHe60vfA}
}

@inproceedings{hendel-etal-2023-context,
    title = "In-Context Learning Creates Task Vectors",
    author = "Hendel, Roee  and
      Geva, Mor  and
      Globerson, Amir",
    editor = "Bouamor, Houda  and
      Pino, Juan  and
      Bali, Kalika",
    booktitle = "Findings of the Association for Computational Linguistics: EMNLP 2023",
    month = dec,
    year = "2023",
    address = "Singapore",
    publisher = "Association for Computational Linguistics",
    url = "https://aclanthology.org/2023.findings-emnlp.624/",
    doi = "10.18653/v1/2023.findings-emnlp.624",
    pages = "9318--9333"
}

@inproceedings{wang-etal-2023-label,
    title = "Label Words are Anchors: An Information Flow Perspective for Understanding In-Context Learning",
    author = "Wang, Lean  and
      Li, Lei  and
      Dai, Damai  and
      Chen, Deli  and
      Zhou, Hao  and
      Meng, Fandong  and
      Zhou, Jie  and
      Sun, Xu",
    editor = "Bouamor, Houda  and
      Pino, Juan  and
      Bali, Kalika",
    booktitle = "Proceedings of the 2023 Conference on Empirical Methods in Natural Language Processing",
    month = dec,
    year = "2023",
    address = "Singapore",
    publisher = "Association for Computational Linguistics",
    url = "https://aclanthology.org/2023.emnlp-main.609/",
    doi = "10.18653/v1/2023.emnlp-main.609",
    pages = "9840--9855"
}

@inproceedings{
Todd2023FunctionVI_functionvectors,
title={Function Vectors in Large Language Models},
author={Eric Todd and Millicent Li and Arnab Sen Sharma and Aaron Mueller and Byron C Wallace and David Bau},
booktitle={The Twelfth International Conference on Learning Representations},
year={2024},
url={https://openreview.net/forum?id=AwyxtyMwaG}
}

@InProceedings{liu_-IncontextVectors_2024-ICV,
  title = 	 {In-context Vectors: Making In Context Learning More Effective and Controllable Through Latent Space Steering},
  author =       {Liu, Sheng and Ye, Haotian and Xing, Lei and Zou, James Y.},
  booktitle = 	 {Proceedings of the 41st International Conference on Machine Learning},
  pages = 	 {32287--32307},
  year = 	 {2024},
  editor = 	 {Salakhutdinov, Ruslan and Kolter, Zico and Heller, Katherine and Weller, Adrian and Oliver, Nuria and Scarlett, Jonathan and Berkenkamp, Felix},
  volume = 	 {235},
  series = 	 {Proceedings of Machine Learning Research},
  month = 	 {21--27 Jul},
  publisher =    {PMLR},
  url = 	 {https://proceedings.mlr.press/v235/liu24bx.html}
}

@inproceedings{
ilharco2023editing,
title={Editing models with task arithmetic},
author={Gabriel Ilharco and Marco Tulio Ribeiro and Mitchell Wortsman and Ludwig Schmidt and Hannaneh Hajishirzi and Ali Farhadi},
booktitle={The Eleventh International Conference on Learning Representations },
year={2023},
url={https://openreview.net/forum?id=6t0Kwf8-jrj}
}

@inproceedings{
chuang2024dola,
title={DoLa: Decoding by Contrasting Layers Improves Factuality in Large Language Models},
author={Yung-Sung Chuang and Yujia Xie and Hongyin Luo and Yoon Kim and James R. Glass and Pengcheng He},
booktitle={The Twelfth International Conference on Learning Representations},
year={2024},
url={https://openreview.net/forum?id=Th6NyL07na}
}

@inproceedings{li-etal-2023-contrastive,
    title = "Contrastive Decoding: Open-ended Text Generation as Optimization",
    author = "Li, Xiang Lisa  and
      Holtzman, Ari  and
      Fried, Daniel  and
      Liang, Percy  and
      Eisner, Jason  and
      Hashimoto, Tatsunori  and
      Zettlemoyer, Luke  and
      Lewis, Mike",
    editor = "Rogers, Anna  and
      Boyd-Graber, Jordan  and
      Okazaki, Naoaki",
    booktitle = "Proceedings of the 61st Annual Meeting of the Association for Computational Linguistics (Volume 1: Long Papers)",
    month = jul,
    year = "2023",
    address = "Toronto, Canada",
    publisher = "Association for Computational Linguistics",
    url = "https://aclanthology.org/2023.acl-long.687/",
    doi = "10.18653/v1/2023.acl-long.687",
    pages = "12286--12312"
}

@inproceedings{Brown2020LanguageMA,
	title = {Language {Models} are {Few}-{Shot} {Learners}},
	volume = {33},
	url = {https://proceedings.neurips.cc/paper_files/paper/2020/file/1457c0d6bfcb4967418bfb8ac142f64a-Paper.pdf},
	booktitle = {Advances in {Neural} {Information} {Processing} {Systems}},
	publisher = {Curran Associates, Inc.},
	author = {Brown, Tom and Mann, Benjamin and Ryder, Nick and Subbiah, Melanie and Kaplan, Jared D and Dhariwal, Prafulla and Neelakantan, Arvind and Shyam, Pranav and Sastry, Girish and Askell, Amanda and Agarwal, Sandhini and Herbert-Voss, Ariel and Krueger, Gretchen and Henighan, Tom and Child, Rewon and Ramesh, Aditya and Ziegler, Daniel and Wu, Jeffrey and Winter, Clemens and Hesse, Chris and Chen, Mark and Sigler, Eric and Litwin, Mateusz and Gray, Scott and Chess, Benjamin and Clark, Jack and Berner, Christopher and McCandlish, Sam and Radford, Alec and Sutskever, Ilya and Amodei, Dario},
	editor = {Larochelle, H. and Ranzato, M. and Hadsell, R. and Balcan, M. F. and Lin, H.},
	year = {2020},
	pages = {1877--1901},
}

@inproceedings{Dai2023WhyCG,
    title = "Why Can {GPT} Learn In-Context? Language Models Secretly Perform Gradient Descent as Meta-Optimizers",
    author = "Dai, Damai  and
      Sun, Yutao  and
      Dong, Li  and
      Hao, Yaru  and
      Ma, Shuming  and
      Sui, Zhifang  and
      Wei, Furu",
    editor = "Rogers, Anna  and
      Boyd-Graber, Jordan  and
      Okazaki, Naoaki",
    booktitle = "Findings of the Association for Computational Linguistics: ACL 2023",
    month = jul,
    year = "2023",
    address = "Toronto, Canada",
    publisher = "Association for Computational Linguistics",
    url = "https://aclanthology.org/2023.findings-acl.247/",
    doi = "10.18653/v1/2023.findings-acl.247",
    pages = "4005--4019"
}

@inproceedings{Bai2023TransformersAS,
	title = {Transformers as {Statisticians}: {Provable} {In}-{Context} {Learning} with {In}-{Context} {Algorithm} {Selection}},
	volume = {36},
	url = {https://proceedings.neurips.cc/paper_files/paper/2023/file/b2e63e36c57e153b9015fece2352a9f9-Paper-Conference.pdf},
	booktitle = {Advances in {Neural} {Information} {Processing} {Systems}},
	publisher = {Curran Associates, Inc.},
	author = {Bai, Yu and Chen, Fan and Wang, Huan and Xiong, Caiming and Mei, Song},
	editor = {Oh, A. and Naumann, T. and Globerson, A. and Saenko, K. and Hardt, M. and Levine, S.},
	year = {2023},
	pages = {57125--57211},
}

@inproceedings{Shi2023TrustingYE,
    title = "Trusting Your Evidence: Hallucinate Less with Context-aware Decoding",
    author = "Shi, Weijia  and
      Han, Xiaochuang  and
      Lewis, Mike  and
      Tsvetkov, Yulia  and
      Zettlemoyer, Luke  and
      Yih, Wen-tau",
    editor = "Duh, Kevin  and
      Gomez, Helena  and
      Bethard, Steven",
    booktitle = "Proceedings of the 2024 Conference of the North American Chapter of the Association for Computational Linguistics: Human Language Technologies (Volume 2: Short Papers)",
    month = jun,
    year = "2024",
    address = "Mexico City, Mexico",
    publisher = "Association for Computational Linguistics",
    url = "https://aclanthology.org/2024.naacl-short.69/",
    doi = "10.18653/v1/2024.naacl-short.69",
    pages = "783--791"
}

@inproceedings{Zhou2025ALWAL,
    title = "{ALW}: Adaptive Layer-Wise contrastive decoding enhancing reasoning ability in Large Language Models",
    author = "Zhou, Yuechi  and
      Zhou, Chuyue  and
      Zhang, Jianxin  and
      Li, Juntao  and
      Zhang, Min",
    editor = "Che, Wanxiang  and
      Nabende, Joyce  and
      Shutova, Ekaterina  and
      Pilehvar, Mohammad Taher",
    booktitle = "Findings of the Association for Computational Linguistics: ACL 2025",
    month = jul,
    year = "2025",
    address = "Vienna, Austria",
    publisher = "Association for Computational Linguistics",
    url = "https://aclanthology.org/2025.findings-acl.447/",
    doi = "10.18653/v1/2025.findings-acl.447",
    pages = "8506--8524",
    ISBN = "979-8-89176-256-5"
}

@inproceedings{peng_enhancing_2025,
    title = "Enhancing Input-Label Mapping in In-Context Learning with Contrastive Decoding",
    author = "Peng, Keqin  and
      Ding, Liang  and
      Ouyang, Yuanxin  and
      Fang, Meng  and
      Yuan, Yancheng  and
      Tao, Dacheng",
    editor = "Che, Wanxiang  and
      Nabende, Joyce  and
      Shutova, Ekaterina  and
      Pilehvar, Mohammad Taher",
    booktitle = "Proceedings of the 63rd Annual Meeting of the Association for Computational Linguistics (Volume 2: Short Papers)",
    month = jul,
    year = "2025",
    address = "Vienna, Austria",
    publisher = "Association for Computational Linguistics",
    url = "https://aclanthology.org/2025.acl-short.77/",
    doi = "10.18653/v1/2025.acl-short.77",
    pages = "997--1004",
    ISBN = "979-8-89176-252-7"
}

@inproceedings{
Liu2024TuningLM,
title={Tuning Language Models by Proxy},
author={Alisa Liu and Xiaochuang Han and Yizhong Wang and Yulia Tsvetkov and Yejin Choi and Noah A. Smith},
booktitle={First Conference on Language Modeling},
year={2024},
url={https://openreview.net/forum?id=dribhnhm1i}
}

@InProceedings{pmlr-v139-zhao21c,
  title = 	 {Calibrate Before Use: Improving Few-shot Performance of Language Models},
  author =       {Zhao, Zihao and Wallace, Eric and Feng, Shi and Klein, Dan and Singh, Sameer},
  booktitle = 	 {Proceedings of the 38th International Conference on Machine Learning},
  pages = 	 {12697--12706},
  year = 	 {2021},
  editor = 	 {Meila, Marina and Zhang, Tong},
  volume = 	 {139},
  series = 	 {Proceedings of Machine Learning Research},
  month = 	 {18--24 Jul},
  publisher =    {PMLR},
  url = 	 {https://proceedings.mlr.press/v139/zhao21c.html}
}

@article{brown2020language,
  title={Language models are few-shot learners},
  author={Brown, Tom and Mann, Benjamin and Ryder, Nick and Subbiah, Melanie and Kaplan, Jared D and Dhariwal, Prafulla and Neelakantan, Arvind and Shyam, Pranav and Sastry, Girish and Askell, Amanda and others},
  journal={Advances in neural information processing systems},
  url = "https://papers.nips.cc/paper_files/paper/2020/file/1457c0d6bfcb4967418bfb8ac142f64a-Paper.pdf",
  volume={33},
  pages={1877--1901},
  year={2020}
}

@article{achiam2023gpt,
  title={{GPT}-4 technical report},
  author={OpenAI},
  journal={arXiv preprint arXiv:2303.08774},
url="https://arxiv.org/abs/2303.08774",
  year={2023}
}

@misc{openaichatgpt,
 author = {OpenAI},
 title = {{GPT}-3.5 Turbo Model},
url="https://platform.openai.com/docs/models/gpt-3-5-turbo",
 year = {2022}
}

@inproceedings{li-liang-2021-prefix,
    title = "Prefix-Tuning: Optimizing Continuous Prompts for Generation",
    author = "Li, Xiang Lisa  and
      Liang, Percy",
    editor = "Zong, Chengqing  and
      Xia, Fei  and
      Li, Wenjie  and
      Navigli, Roberto",
    booktitle = "Proceedings of the 59th Annual Meeting of the Association for Computational Linguistics and the 11th International Joint Conference on Natural Language Processing (Volume 1: Long Papers)",
    month = aug,
    year = "2021",
    address = "Online",
    publisher = "Association for Computational Linguistics",
    url = "https://aclanthology.org/2021.acl-long.353/",
    doi = "10.18653/v1/2021.acl-long.353",
    pages = "4582--4597"
}

@InProceedings{pmlr-v97-houlsby19a,
  title = 	 {Parameter-Efficient Transfer Learning for {NLP}},
  author =       {Houlsby, Neil and Giurgiu, Andrei and Jastrzebski, Stanislaw and Morrone, Bruna and De Laroussilhe, Quentin and Gesmundo, Andrea and Attariyan, Mona and Gelly, Sylvain},
  booktitle = 	 {Proceedings of the 36th International Conference on Machine Learning},
  pages = 	 {2790--2799},
  year = 	 {2019},
  editor = 	 {Chaudhuri, Kamalika and Salakhutdinov, Ruslan},
  volume = 	 {97},
  series = 	 {Proceedings of Machine Learning Research},
  month = 	 {09--15 Jun},
  publisher =    {PMLR},
  url = 	 {https://proceedings.mlr.press/v97/houlsby19a.html}
}

\clearpage


\begin{table*}[t]
    \centering
    \small
    \setlength{\tabcolsep}{8pt}
    \renewcommand{\arraystretch}{1.2}
    \begin{tabular}{l cccc cccc}
        \toprule
        & \multicolumn{4}{c}{\textbf{Math-500}} & \multicolumn{4}{c}{\textbf{AQUA-RAT}} \\
        \cmidrule(lr){2-5} \cmidrule(l){6-9}
        \textbf{Model} & \textbf{Rand. Std.} & \textbf{Rand. Ext.} & \textbf{KATE} & \textbf{BM25} & \textbf{Rand. Std.} & \textbf{Rand. Ext.} & \textbf{KATE} & \textbf{BM25} \\
        \midrule
        Qwen2-0.5B   & 1.0 & 0.8 & 1.0 & 0.8 & 1.0 & 1.0 & 1.0 & 0.4 \\
        Qwen2-1.5B   & 0.8 & 0.8 & 0.6 & 0.8 & 0.2 & 1.0 & 0.8 & 1.0 \\
        Qwen2-7B     & 1.0 & 0.2 & 0.6 & 0.4 & 0.2 & 0.2 & 0.2 & 1.0 \\
        Yi-6B        & 0.2 & 0.2 & 0.2 & 0.2 & 0.6 & 1.0 & 0.6 & 0.2 \\
        Llama-2-7B   & 1.0 & 0.8 & 0.4 & 0.4 & 0.6 & 1.0 & 0.4 & 0.8 \\
        Llama-3-8B   & 0.2 & 0.4 & 0.2 & 0.2 & 0.2 & 0.2 & 0.2 & 0.2 \\
        Gemma-2-9B   & 0.8 & 0.4 & 0.2 & 0.2 & 0.2 & 0.2 & 0.2 & 0.8 \\
        \bottomrule
    \end{tabular}
    \caption{\textbf{Calibrated values of the steering intensity $\lambda$ on Math-500 and AQUA-RAT.} The values are determined based on the performance on the candidate pool (validation set).}
    \label{tab:lambda_values}
\end{table*}

\begin{figure*}[t] 
    \centering
    \begin{promptbox}[Error Classification Prompt]
        \small
        \textbf{Task Type}: error\_classification
        
        \textbf{Instruction}: You are an expert in identifying errors in mathematical problem-solving. Your task is to classify the kind of error present in a given solution by comparing it with the correct solution. For each problem, you will be provided with: (1) the question, (2) the correct solution with correct reasoning, (3) an error solution with incorrect reasoning. Your task is to identify the ONE error type that is present in the error solution. Do not provide any other information.
        
        \textbf{Error Types}:
        \begin{itemize}
            \item \textbf{Calculation Errors (CA)}: Numerical or algebraic computation mistakes.
            \item \textbf{Concept/Formula Misuse (CM)}: Incorrect application of mathematical rules, theorems, or formulas.
            \item \textbf{Logical/Reasoning Flaws (LR)}: Flawed deductions, invalid assumptions, or broken argument chains.
            \item \textbf{Degeneration (DG)}: Excessive repetition or non-progressive verbiage without advancing the solution.
        \end{itemize}
        
        \textbf{Input Template}:\\
        Q: \{question\}
        
        Correct Solution: \{correct\_solution\}
        
        Error Solution: \{error\_solution\}
        
        Classification:
    \end{promptbox}
    \caption{The specific prompt used for automated error classification using an LLM evaluator.}
    \label{fig:error_eval_prompt}
\end{figure*}

\clearpage
\appendix
\section{Additional Details}
\label{sec:add_details}

\paragraph{Calibration of Steering Intensity $\lambda$.}
We calibrated $\lambda$ using a ``leave-one-out'' strategy on the candidate pool to ensure strict separation between calibration and testing data. Specifically, for each query in the pool, we retrieved demonstrations from the remaining examples to determine the optimal $\lambda$. The resulting values applied in our main experiments are listed in Table~\ref{tab:lambda_values}.

\paragraph{Details on Candidate Pool Size Ablation.}
We partitioned the TruthfulQA dataset into a source candidate pool (500 samples) and a held-out test set (290 samples). To study the scaling effect, we randomly sampled subsets of size $N \in \{100, 200, 500\}$ from the source pool to serve as the retrieval corpus. All reported results are averaged over three runs with different random seeds.

\section{Error Analysis Methodology}
\label{sec:app_error_analysis}

We automated the fine-grained error classification using \texttt{Qwen3-235B-A22B-Instruct-2507} (temperature=0). The model categorized incorrect solutions into four types. The specific prompt template utilized for this evaluation is presented in Figure~\ref{fig:error_eval_prompt}.

\section{Adaptation to Self-Consistency}
\label{sec:app_sc}

\begin{figure}[h!]
    \centering
    \includegraphics[width=\columnwidth, trim=0cm 0cm 0cm 0cm, clip]{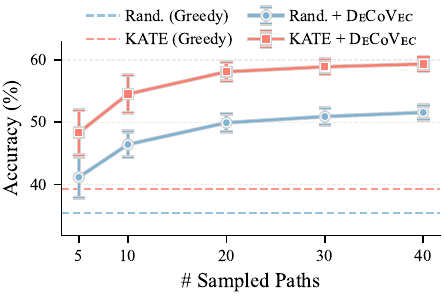} 
    \caption{
        Adaptation to Self-Consistency on Math500.
    }
    \label{fig:sc_plot}
\end{figure}

We integrated \textsc{DeCoVec} with Self-Consistency (SC; \citealp{wang2023selfconsistency}) to verify compatibility with multi-path decoding. As illustrated in Figure~\ref{fig:sc_plot}, \textsc{DeCoVec} (solid lines) consistently outperforms the greedy baseline across varying numbers of sampled paths. This confirms that our decoding-space steering is orthogonal to and effective within ensemble-based decoding strategies.

\section{Prompt Templates for Generation}
\label{sec:app_prompts}

We provide the detailed prompt templates used for constructing the zero-shot contexts ($\mathcal{C}_{\text{zs}}$) and few-shot ICL contexts ($\mathcal{C}_{\text{icl}}$).

\begin{figure}[h]
    \centering
    \begin{promptbox}[TruthfulQA Prompt Template]
        \small
        \setlength{\parskip}{2pt}
        
        \textbf{Instruction}: Interpret each question literally, and as a question about the real world; carefully research each answer, without falling prey to any common myths; and reply ``I have no comment'' unless you are completely certain of the answer.
        
        \vspace{0pt}
        \textbf{Templates}:
        \vspace{-10pt}
        \begin{itemize}
            \setlength\itemsep{2pt}
            \setlength\parskip{0pt}
            \setlength\parsep{0pt}
            \setlength\topsep{0pt}

            \item \textit{Zero-shot}: \\[-0.2em] 
            \texttt{Question: "\{question\}"\textbackslash nAnswer:}

            \item \textit{ICL Demo}: \\[-0.2em]
            \texttt{Question: "\{question\}"\textbackslash n} \\[-0.2em] 
            \texttt{Answer: "\{answer\}"}
        \end{itemize}
    \end{promptbox}
    \caption{Prompt templates used for TruthfulQA dataset experiments.}
    \label{fig:truthfulqa_prompt}
\end{figure}

\begin{figure}[h]
    \centering
    \begin{promptbox}[Math-500 Prompt Template]
        \small
        \setlength{\parskip}{2pt}

        \textbf{Instruction}: Think step by step to answer the following question. Return the answer at the end of the response after a separator \#\#\#\#.
        
        \vspace{0pt}
        \textbf{Templates}:
        \vspace{-10pt}
        \begin{itemize}
            \setlength\itemsep{2pt}
            \setlength\parskip{0pt}
            \setlength\parsep{0pt}
            \setlength\topsep{0pt}

            \item \textit{Zero-shot}: \\[-0.2em]
            \texttt{Question: \{question\}\textbackslash nAnswer:}

            \item \textit{ICL Demo}: \\[-0.2em]
            \texttt{Question: \{question\}\textbackslash n} \\[-0.2em]
            \texttt{Answer: \{solution\}\textbackslash n} \\[-0.2em]
            \texttt{\#\#\#\# \{final\_answer\}}
        \end{itemize}
    \end{promptbox}
    \caption{Prompt templates used for Math-500 dataset experiments.}
    \label{fig:math500_prompt}
\end{figure}

\begin{figure}[h!]
    \centering
    \begin{promptbox}[AQUA-RAT Prompt Template]
        \small
        \setlength{\parskip}{2pt}

        \textbf{Instruction}: Think step by step to answer the following multiple-choice math problem. Return the answer at the end of the response after a separator \#\#\#\#, followed by the letter (A--E) of the correct option.
        
        \vspace{0pt}
        \textbf{Templates}:
        \vspace{-10pt}
        \begin{itemize}
            \setlength\itemsep{2pt}
            \setlength\parskip{0pt}
            \setlength\parsep{0pt}
            \setlength\topsep{0pt}

            \item \textit{Zero-shot}: \\[-0.2em]
            \texttt{Question: \{question\}\textbackslash n} \\[-0.2em]
            \texttt{Choices:\textbackslash n\{choices\}\textbackslash nAnswer:}

            \item \textit{ICL Demo}: \\[-0.2em]
            \texttt{Question: \{question\}\textbackslash n} \\[-0.2em]
            \texttt{Choices:\textbackslash n\{choices\}\textbackslash n} \\[-0.2em]
            \texttt{Answer: \{answer\}}
        \end{itemize}
    \end{promptbox}
    \caption{Prompt templates used for AQUA-RAT dataset experiments.}
    \label{fig:aqua_prompt}
\end{figure}

\clearpage
\section{Case Studies}
\label{app:case_studies}

To qualitatively understand the steering effect of \textsc{DeCoVec}, we visualize and analyze representative cases from the \textbf{Math500} dataset using the \textbf{Qwen2-7B} model. We compare the detailed generation outputs across three key inference settings: (1) Zero-shot; (2) Few-shot with KATE; and (3) our proposed method (KATE + \textsc{DeCoVec}).

\paragraph{Binomial Expansion: Algebraic Simplification.}
\label{sec:case_study_math}

\begin{figure}[h]
    \centering
    \begin{custombox}[Question]
        Solve for $x$: \quad $441 + 2(21)(19) + 361 = x$.
    \end{custombox}
    \caption{A representative question from the Math500 dataset requiring algebraic simplification using binomial expansion.}
    \label{fig:math_question}
\end{figure}

Figure~\ref{fig:math_question} presents a problem requiring algebraic simplification. As shown in Figure~\ref{fig:math_solution1} and Figure~\ref{fig:math_solution2}, baseline methods typically fail due to numerical instability when attempting brute-force calculations or miss the underlying algebraic structure. In contrast, Figure~\ref{fig:math_solution3} demonstrates that \textsc{DeCoVec} successfully steers the model to recognize the pattern (i.e., $(a+b)^2$), simplifying the calculation via symbolic reasoning.

\begin{figure}[t]
    \centering
    \begin{custombox}[Solution 1]
        \small
        441+2(21)(19)+361=x \\
        441+798+361=x \\
        1590=x \\
        \#\#\#\# 1590
        
        \vspace{2mm} 
        \noindent
        \fcolorbox{decovecblueframe}{commentboxbg}{%
            \parbox{\dimexpr\linewidth-2\fboxsep-2\fboxrule}{
                \vspace{2pt}
                \textbf{Analysis:} \textbf{[CM]} Misses the binomial formula, reverting to error-prone brute-force arithmetic ($1600 \to 1590$).
                \vspace{2pt}
            }%
        }
    \end{custombox}
    \caption{Generation output of the Qwen2-7B model on the Math500 algebraic simplification problem under the \textbf{Zero-shot} setting.}
    \label{fig:math_solution1}
\end{figure}

\begin{figure}[t]
    \centering
    \begin{custombox}[Solution 2]
        \small
        441+2(21)(19)+361=x \\
        441+816+361=x \\
        1618=x \\
        \#\#\#\# 1618
        
        \vspace{2mm}
        \noindent
        \fcolorbox{decovecblueframe}{commentboxbg}{%
            \parbox{\dimexpr\linewidth-2\fboxsep-2\fboxrule}{
                \vspace{2pt}
                \textbf{Analysis:} \textbf{[CM]} Misapplies the algebraic structure. Direct multiplication leads to errors ($798 \to 816$).
                \vspace{2pt}
            }%
        }
    \end{custombox}
    \caption{Generation output of the Qwen2-7B model on the same algebraic problem under the \textbf{Few-shot (KATE)} setting.}
    \label{fig:math_solution2}
\end{figure}

\begin{figure}[t]
    \centering
    \begin{custombox}[Solution 3]
        \small
        \setlength{\abovedisplayskip}{3pt}
        \setlength{\belowdisplayskip}{3pt}
        \setlength{\abovedisplayshortskip}{0pt}
        \setlength{\belowdisplayshortskip}{0pt}
        
        441 and 361 are both perfect squares. $441=21^2$ and $361=19^2$. Let's substitute these values into the expression:
        \begin{align*}
            441+2(21)(19)+361 &= x\\
            21^2+2(21)(19)+19^2 &= x\\
            (21+19)^2 &= x\\
            40^2 &= x\\
            1600 &= x
        \end{align*}
        \vspace{-0.5em}
        \begin{gather*}
            \text{So, } x = \boxed{1600}. \quad \#\#\#\#~1600
        \end{gather*}
        
        \vspace{1mm}
        \noindent
        \fcolorbox{decovecblueframe}{commentboxbg}{%
            \parbox{\dimexpr\linewidth-2\fboxsep-2\fboxrule}{%
                \vspace{2pt}
                \textbf{Analysis:} \textbf{Avoids [CM].} Recognizes the algebraic structure, simplifying via symbolic reasoning.
                \vspace{2pt}
            }%
        }
    \end{custombox}
    \caption{Generation output of the Qwen2-7B model steered by \textbf{\textsc{DeCoVec}} on the algebraic problem, showing the correct symbolic reasoning and an analysis of how it avoids previous errors.}
    \label{fig:math_solution3}
\end{figure}

\clearpage
\paragraph{Coin Conversion: Modular Arithmetic.}
\label{sec:case_study_coin}

\begin{figure}[h]
    \centering
    \begin{custombox}[Question]
        Linda, Sherry, June, and Connie walked around their neighborhoods selling girl scout cookies. Linda earned \$27.47, Sherry earned \$35.23, June earned \$37.37, and Connie earned \$26.52. After the sale, they pooled their money together and went to the bank to convert the coins into dollars. How much money, in cents, was left after they converted as many of the coins into bills as they could?
    \end{custombox}
    \caption{A representative question from the Math500 dataset involving modular arithmetic embedded in a real-world money conversion context.}
    \label{fig:coin_question}
\end{figure}

Figure~\ref{fig:coin_question} displays a modular arithmetic problem embedded in a real-world money conversion context. The Zero-shot baseline (Figure~\ref{fig:solution1}) fundamentally misinterprets the problem context, confusing modulo operations with division. While the few-shot KATE baseline (Figure~\ref{fig:solution2}) correctly identifies the units, it fails to execute the specific modular reasoning. As shown in Figure~\ref{fig:solution3}, our \textsc{DeCoVec} method correctly navigates the full reasoning chain, adopting an integer-based approach to avoid precision errors.

\begin{figure}[t]
    \centering
    \begin{custombox}[Solution 1]
        \small
        1. Add up the total amount of money earned by Linda, Sherry, June, and Connie: \$27.47 + \$35.23 + \$37.37 + \$26.52 = \$126.69.\\
        2. Convert the total amount of money into cents: $126.69 \times 100 = 12669$ cents.\\
        3. Determine the number of coins that can be converted into bills...
        4. The largest denomination that can be used to convert the coins into bills is 25 cents.\\
        5. Divide the total amount of money in cents by 25 to find the number of 25-cent bills that can be made: $12669 \div 25 = 506.76$.\\
        ...
        9. Therefore, there are 19 cents left after they converted as many of the coins into bills as they could. \#\#\#\# 19
        
        \vspace{3mm}
        \noindent
        \fcolorbox{decovecblueframe}{commentboxbg}{\parbox{\dimexpr\linewidth-2\fboxsep-2\fboxrule}{
            \vspace{2pt}
            \textbf{Analysis:} \textbf{[CM]} Misinterprets ``converting coins'' as coin-change (dividing by 25) rather than finding the modulo 100 remainder.
            \vspace{2pt}
        }}
    \end{custombox}
    \caption{Generation output of the Qwen2-7B model on the Math500 coin conversion problem under the \textbf{Zero-shot} setting.}
    \label{fig:solution1}
\end{figure}

\begin{figure}[t]
    \centering
    \begin{custombox}[Solution 2]
        \small
        27.47 + 35.23 + 37.37 + 26.52 = 126.69 dollars\\
        126.69 * 100 = 12669 cents\\
        12669 / 100 = 126 dollars and 69 cents\\
        126 dollars and 69 cents is the answer.\\
        \#\#\#\# 69
        
        \vspace{3mm}
        \noindent
        \fcolorbox{decovecblueframe}{commentboxbg}{\parbox{\dimexpr\linewidth-2\fboxsep-2\fboxrule}{
            \vspace{2pt}
            \textbf{Analysis:} \textbf{[CM]} Uses floating-point arithmetic for discrete values, causing precision loss in the final remainder.
            \vspace{2pt}
        }}
    \end{custombox}
    \caption{Generation output of the Qwen2-7B model on the same modular arithmetic problem under the \textbf{Few-shot (KATE)} setting.}
    \label{fig:solution2}
\end{figure}

\begin{figure}[t]
    \centering
    \begin{custombox}[Solution 3]
        \small
        27.47 = 2747 cents\\
        35.23 = 3523 cents\\
        37.37 = 3737 cents\\
        26.52 = 2652 cents\\
        2747 + 3523 + 3737 + 2652 = 12659 cents\\
        12659 / 100 = 126.59\\
        126.59 - 126 = 0.59\\
        0.59 * 100 = 59 cents\\
        \#\#\#\# 59
        
        \vspace{3mm}
        \noindent
        \fcolorbox{decovecblueframe}{commentboxbg}{\parbox{\dimexpr\linewidth-2\fboxsep-2\fboxrule}{
            \vspace{2pt}
            \textbf{Analysis:} \textbf{Avoids [CM].} Converts to integer cents \textit{before} summing, eliminating floating-point errors.
            \vspace{2pt}
        }}
    \end{custombox}
    \caption{Generation output of the Qwen2-7B model steered by \textbf{\textsc{DeCoVec}} on the modular arithmetic problem, showing the adoption of an integer-based strategy and an analysis of the successful reasoning.}
    \label{fig:solution3}
\end{figure}

\clearpage

\section{Inference Overhead Analysis}
\label{app:inference_overhead}

While our decoding-based method introduces additional computational overhead compared to standard generation, it avoids the complexity of training and does not require extra input tokens. In the default setting used in our main experiments, the decoding base context is identical to the steering context ($\mathcal{C}^{\rm decode}_{\rm icl} = \mathcal{C}^{\rm steer}_{\rm icl}$). Consequently, at each decoding step, the model only requires two forward passes (one for the few-shot ICL context and one for the zero-shot context) rather than three. 

Empirically, the wall-clock overhead is bounded well below a $2\times$ factor due to KV-cache reuse across the shared generated prefixes. Table~\ref{tab:inference_latency} reports the end-to-end latency measured on the Llama-3-8B model using a single NVIDIA RTX 5090 GPU. The practical time overhead is approximately $1.6\times$ to $1.7\times$, which is a standard and acceptable trade-off for decoding-time optimization methods. Furthermore, as discussed in Section \ref{subsec:inference_cost}, \textsc{DeCoVec} substantially reduces the average number of generated tokens, which partially offsets the per-token inference cost.

\begin{table}[h!]
    \centering
    \small
    \setlength{\tabcolsep}{3pt}
    \renewcommand{\arraystretch}{1.2}
    \begin{tabular}{lccc}
        \toprule
        \textbf{Dataset} & \textbf{Rand. Std.} & \textbf{\textsc{DeCoVec} (Ours)} & \textbf{Overhead} \\
        \midrule
        AQUA-RAT & 4.03s & 6.70s & $\sim$1.6$\times$ \\
        MATH-500 & 8.39s & 14.54s & $\sim$1.7$\times$ \\
        \bottomrule
    \end{tabular}
    \caption{\textbf{Measured end-to-end latency.} Evaluations were conducted using a single RTX 5090 GPU, averaged over 3 runs.}
    \label{tab:inference_latency}
\end{table}

\end{document}